\documentclass[10pt,journal,compsoc]{IEEEtran}
\ifCLASSOPTIONcompsoc
  \usepackage[nocompress]{cite}
\else
  \usepackage{cite}
\fi
\ifCLASSINFOpdf
\usepackage[pdftex]{graphicx}
\graphicspath{{images/}}
\DeclareGraphicsExtensions{.pdf,.jpeg,.png,.tif}
\else
\usepackage[dvips]{graphicx}
\graphicspath{{images/}}
\DeclareGraphicsExtensions{.eps}
\fi

\usepackage[cmex10]{amsmath}
\usepackage{bbm}

\ifCLASSOPTIONcompsoc
\usepackage[caption=false,font=normalsize,labelfont=sf,textfont=sf]{subfig}
\else
\usepackage[caption=false,font=footnotesize]{subfig}
\fi
\usepackage{stfloats}

\usepackage{pgfplots}
\usepgfplotslibrary{colormaps}
\usetikzlibrary{shapes}
\usepackage{cite}
\usepackage{siunitx}

\newcommand{\kernel}{\phi}

\usepackage{amsmath}
\usepackage{amssymb}
\usepackage{wasysym}

\usepackage{mathtools} % for \mathclap to reduce spacing

\usepackage{listings}
\usepackage{pdfpages}
\usepackage{multirow}
\usepackage{lineno}
\usepackage{rotating}

\usepackage{graphicx}
\usepackage{hyperref}
\usepackage[english]{babel}
\usepackage[utf8]{inputenc}

\usepackage{algorithm}
\usepackage{algorithmic}

\usepackage{tikz}
\usetikzlibrary{positioning,calc,decorations.text,patterns}

\usepackage{pgfplots}
\usepackage{pgfplotstable}
\pgfplotsset{compat=newest}
\pgfplotsset{plot coordinates/math parser=false}
\newlength\figureheight
\newlength\figurewidth 

\newcommand\inner[2]{\left\langle #1, #2 \right\rangle}

\usepackage{ifthen}
\usepackage{etoolbox}
\newcommand\corrected[2][]{#2}
\newcommand\restructured[2][]{#2}
\newcommand\todo[2][]{\textcolor{red}{\ifthenelse{\equal{#1}{}}{}{[#1] }#2}}
\newcommand\correctedfinal[2][]{#2}
\begin{document}
\title{Kernel-Based Generalized Median Computation for Consensus Learning}

\author{
Andreas Nienk\"otter,
Xiaoyi Jiang,~\IEEEmembership{Senior Member,~IEEE}
 % stops a space
\IEEEcompsocitemizethanks{\IEEEcompsocthanksitem A. Nienk\"otter and X. Jiang are with the Faculty of Mathematics and Computer Science, University of M\"unster, M\"unster, Germany.}
%\protect\\
%E-mail: see http://www.michaelshell.org/contact.html
}


\markboth{Accepted for IEEE Transactions on Pattern Analysis and Machine Intelligence, 2022}%
{Shell \MakeLowercase{\textit{et al.}}: Bare Demo of IEEEtran.cls for Computer Society Journals}



\IEEEtitleabstractindextext{%
\begin{abstract}
Computing a consensus object from a set of given objects is a core problem in machine learning and pattern recognition. One popular approach is to formulate it as an optimization problem using the generalized median. Previous methods like the Prototype and Distance-Preserving Embedding methods transform objects into a vector space, solve the generalized median problem in this space, and inversely transform back into the original space. Both of these methods have been successfully applied to a wide range of object domains, where the generalized median problem has inherent high computational complexity (typically $\mathcal{NP}$-hard) and therefore approximate solutions are required. Previously, explicit embedding methods were used in the computation, which often do not reflect the spatial relationship between objects exactly. In this work we introduce a kernel-based generalized median framework that is applicable to both positive definite and indefinite kernels. This framework computes the relationship between objects and its generalized median in kernel space, without the need of an explicit embedding. We show that the spatial relationship between objects is more accurately represented in kernel space than in an explicit vector space using easy-to-compute kernels, and demonstrate superior performance of generalized median computation on datasets of three different domains. A software toolbox resulting from our work is made publicly available to encourage other researchers to explore the generalized median computation and applications.
\end{abstract}

}

\maketitle

\IEEEdisplaynontitleabstractindextext

% \fi
\IEEEpeerreviewmaketitle

\IEEEraisesectionheading{\section{Introduction}\label{sec:introduction}}


\IEEEPARstart{O}{ne} commonly used approach to consensus learning is to formulate it as an optimization problem in terms of generalized median computation \cite{nienkoetter2019dpe}. Given a set of objects $O = \{o_1,...,o_n\}$ in domain $\mathcal{O}$ with a distance function $\delta(o_i,o_j)$, the \textit{generalized median} can be expressed as
\begin{align}
\bar{o} = \arg \min_{o \in \mathcal{O}} \underbrace{\sum_{o_i \in O} \delta(o_i, o)}_{SOD(o)}
\label{eq:median}
\end{align}
In other words, the generalized median is an object that has the smallest sum of distances (SOD, also called \textit{consensus error} \cite{Gudfield1997}) to all objects in the input set. Note that the median object is not necessarily part of set $O$. The generalized median is the formalization of the intuitive averaging. As a simple example, when dealing with real numbers, it corresponds to well-known concepts from statistics. In case of $\delta(p,q) = (p-q)^2, \ p,q \in \mathbb{R}$, the generalized median is simply the arithmetic average of the given numbers. Changing to another distance function $\delta(p,q) = |p-q|$ results in the usual median of numbers.

The generalized median is a general concept and has been studied for numerous problem domains related to a broad range of applications. Examples include rankings \cite{Boulakia_2011}, phase or orientation data \cite{Rothaus_2006,Storath_2018}, 3D rigid structures \cite{Ding_NIPS2013}, 3D rotations \cite{Rotation_Averaging_2013,Lee20}, clusterings \cite{Clustering_Ensemble_Survey_2011,Carpineto_2012,Boongoen18}, point-sets \cite{Ding14}, shapes \cite{Berkels_2010}, 3D surfaces \cite{wu_2015}, image segmentation \cite{Franek2010}, sequence data (strings) \cite{jiang2012generalized,nienkoetter2018sequence}, graphs \cite{jiang2001median,mukherjee2009generalized}, altas construction \cite{atlas_2013}, and Grassmann average \cite{Hauberg14}. It is often known under different names in applications, where a specific distance is common, for example geometric median in case of Euclidean vector spaces with the Euclidean distance \cite{ferrer2010generalized}, Steiner string for median strings \cite{Gudfield1997}, Karcher mean for positive definite matrices \cite{lim2012matrix} \restructured{or Kemeny consensus for rankings \cite{tang_mallows_2019}}.

Although being simple in its definition, the optimization task \eqref{eq:median} turns out to be very complex in many domains. Since the median object is not necessary part of the original set, one has to construct a new object from the whole domain space that minimizes the SOD in equation (\ref{eq:median}). Even for the simple string edit distance, it has been proven to be $\mathcal{NP}$-hard \cite{de2000topology}. The same applies to rankings \cite{Boulakia_2011}, ensemble clustering \cite{Krivanek_1986}, median graphs \cite{ferrer2010generalized} and signed permutations \cite{bader2011transposition} using common distance functions, just to name a few examples. Also for the seemingly simple case of $\mathbb{R}^d$ with the Euclidean distance there is no known polynomial-time algorithm, and it is not even known if this problem is in $\mathcal{NP}$ \cite{hakimi2000location}.

Given the high computational complexity, approximate solutions are required to calculate the generalized median in reasonable time. The perturbation strategy in \cite{Carpineto_2012} for ensemble clustering, for instance, starts with an initial clustering and iteratively moves a single object to a different (possibly empty) cluster until the optimization SOD function associated with the newly created clustering increases. More sophisticated methods include metaheuristic strategies,  e.g. genetic algorithms \cite{jiang2001median} that optimize a set of candidate solutions by combining and mutating them, simulated annealing \cite{Carpineto_2010} that optimizes a single candidate by probabilistic optimization, or block coordinate descent \cite{boria2019generalized} minimizing a matrix or vector representation of the generalized median approximation. These methods are typically heavily tailored to a specific domain $\mathcal{O}$ and use intricate domain knowledge in their computation.

The focus of our work lies in domain-independent methods that can compute generalized median solutions for any space. There exist only very few such frameworks. The framework for generalized median approximation in \cite{Franek_SSPR_2012} is motivated by a lower bound \cite{jiang2001median} for generalized median computation in metric spaces. Recently, the prototype-embedding approach \cite{ferrer2010generalized} has received considerable attention to successfully solve a number of $\mathcal{NP}$-hard consensus learning problem instances (strings, graphs, clusterings, biclusterings) with superior quality \cite{nienkoetter2019dpe,jiang2012generalized,ferrer2010generalized,ferrer2011generic,franek2014ensemble}.
In this framework the objects are first embedded into a vector space, where the median computation is much easier than in the general case. The median vector is then transformed back into the original problem space (reconstruction), resulting in an approximation of the generalized median. Further improvement of the prototype-embedding approach includes alternative object reconstruction methods \cite{nienkoetter2016reconstruction} and more accurate distance-preserving embedding methods compared to prototype embedding \cite{nienkoetter2019dpe}.

\corrected[2.3]{This approach can also be applied to structured prediction, which is a broad research topic and has been studied in many domains. In \cite{bekoulis_reconstructing_2017}, for instance, a tree structure of a house plan is predicted from an advertising text by first labelling the text, computing the intra-label dependencies and finally constructing a tree from these dependencies. Other examples include label ranking prediction \cite{korba_structured_2018}, human activity recognition \cite{arzani_switching_2021}, and person re-identification \cite{liao_person_2019}. However, these methods and their embeddings (2D dependency edges in the example) are typically highly specialized to the problem at hand. In contrast, the distance embedding approach is a general method applicable to any domain a distance is known.}

The popular prototype-embedding approach \cite{ferrer2010generalized,nienkoetter2019dpe} is based on \textit{explicit} transformation for the embedding purpose. In this work we propose a novel kernel approach to embedding-based generalized median computation by using an \textit{implicit} transformation in terms of kernel functions. We show that it is possible to handle the generalized median computation \textit{without} knowing the dimension of the embedded space and the concrete embedding.
%introduce a novel kernel based method for the computation of the generalized median of any object, given only a kernel function and and a so called weighted mean function. This method computes the spacial relationship of the objects in a set and the generalized median in the implicit vector space of the kernel function, and uses this information to reconstruct an approximate median object from the these objects.
We evaluate our method on artificial and real datasets of three different problem domains (strings, clusterings, rankings). Several kernel functions will be studied, in particular some of them have the nice property of provably preserving the pairwise distances after embedding in the implicit vector space, which is a highly desired property for embedding-based generalized median computation \cite{nienkoetter2019dpe}. Using these easy-to-compute kernel functions we demonstrate superior performance in terms of the computed median quality compared to the prototype-based explicit embedding method.

The remainder of the paper is structured as follows. In Section \ref{sec:probability}, we give a statistical interpretation of generalized median \eqref{eq:median} as a maximum-likelihood estimator. The prototype-based embedding framework is summarized in Section \ref{sec:history}. Its inherent drawbacks are discussed to motivate our current work. Section \ref{sec:kernel-mean} presents our kernel-based approach to generalized median computation. The experimental evaluation is given in Section \ref{sec:evaluation}. A related software toolbox for public use will be described in Section \ref{sec:toolbox}. Finally, some discussions in Section \ref{sec:conclusion} conclude the paper.

%%%%%%%%%%%%%%%%%%%%%%%%%%%%%%%%%%%%%%%%%%%%%%%%%%%%%%%%%%%%%%%%%%%%%%%%%%%%%
\section{Generalized Median as Maximum-Likelihood Estimator}
\label{sec:probability}

We start with the simple case of 1D signal. If the true, but unknown, signal value is $\bar{x} \in \mathbb{R}$, the measured signal $x \in \mathbb{R}$ is disturbed by an error $x-\bar{x}$. Then, the probability of a measurement with this error can be expressed by an exponential function of the error. Assuming the probability is quadratic in the error, this leads to the common Gaussian distribution
\begin{align*}
	\mathcal{N}(x|\bar{x},\sigma) = \frac{1}{\sqrt{2 \pi} \sigma} \exp \left( -\frac{(x-\bar{x})^2}{2\sigma^2} \right)
\end{align*}
Assuming the error is not quadratic, one can express the probability as the so-called Laplace distribution
\begin{align*}
	\mathcal{L}(x|\bar{x},\sigma) = \frac{1}{2 \sigma} \exp \left( -\frac{|x-\bar{x}|}{\sigma} \right)
\end{align*}
These probability distributions are also known as the second and first law of error, and were first proposed by Laplace around 1775 \cite{wilson1923first}. Although the Gaussian distribution is the more popular one, it has been shown that it does not always reflect the error distribution of natural data, especially if outliers are present \cite{wilson1923first,kotz2012laplace}. Naturally, the Laplace distribution appears for example in the error of navigational data \cite{hsu1979long} or financial data \cite{kotz2012laplace}.

%\corrected{One interesting property is that the Laplace distribution can be expressed as a scale mixture of normal distributions as
%%
%$$\mathcal{L}(x|\bar{x},\sigma) \ \sim \ \bar{x} + \frac{\sigma}{\sqrt{2}} \sqrt{\mathcal{E}(x|1)} \  \mathcal{N}(x|0,1)$$
%%
%using an exponential distribution $\mathcal{E}(x|\lambda)$ \cite{kotz2012laplace}. As such, it can be seen as a mixture of normally distributed values with different variances.}

%\begin{figure}
%  \begin{center}
%	\begin{tikzpicture}[scale=0.6]
%	\begin{axis}[
%	xlabel=$x$,
%	ylabel=$f(x)$,
%	axis lines=middle,
%	xmin = -6,
%	xmax = 6,
%	ytick = {0.5}]
%	\addplot[mark=none] {1/2 * exp(-abs(x-0)/2)};
%	\end{axis}
%	\end{tikzpicture}
%	\end{center}
%	\caption{Laplace distribution $\mathcal{L}(x|\mu,\sigma)$ with $\mu = 0$, $\sigma = 1$.}
	\label{fig:laplace}
%\end{figure}

Now we consider an arbitrary space $\mathcal{O}$. Assuming the true, but unknown, object $\bar{o}$, we make observations $o$ and model the error by the distance $\delta(o, \bar{o})$ in analogy of $|x-\bar{x}|$ in the Laplace distribution for $\mathbb{R}$. Then, one can compute the probability for the occurrence of any object $o$ by
\begin{align*}
\mathcal{L}(o|\bar{o},\sigma) = \frac{1}{2 \sigma} \exp \left( -\frac{\delta(o,\bar{o})}{\sigma} \right)
\end{align*}
Using this distribution, one can estimate the optimal object $\bar{o}$ and the distribution factor $\sigma$ for a set of objects $O = \{o_1,...,o_n\}$ by maximizing the likelihood function
\begin{align*}
	 \nonumber L(\bar{o},\sigma | O) &= \prod_{i=1}^n \frac{1}{2 \sigma} \exp \left( -\frac{\delta(o_i,\bar{o})}{\sigma} \right) \\
	&= \left(\frac{1}{2 \sigma}\right)^n \exp \left(- \frac{\sum_{i=1}^n \delta(o_i,\bar{o})}{\sigma}\right)
\end{align*}
under the assumption that $o_i$ are independent and identically distributed.
This is equivalent to maximizing the logarithm of the likelihood function
\begin{align*}
	\log(L(\bar{o},\sigma | O)) &=  n \log\left(\frac{1}{2 \sigma}\right) - \frac{\sum_{i=1}^n \delta(o_i,\bar{o})}{\sigma}
\end{align*}
Since $\sigma$ and $n$ are constant, an optimal estimation $\bar{o}$ is equal to minimizing the second term (without the negative sign), which corresponds to the definition of generalized median in \eqref{eq:median}. In a next step the optimal $\sigma$ can be computed as
\begin{align*}
\sigma &= \frac{\sum_{i=1}^n \delta(o_i,\bar{o})}{n}
\end{align*}
by setting $\frac{\partial \log(L(\bar{o},\sigma | O))}{\partial \sigma}$ to be zero, which is the normalized SOD.

Therefore, the generalized median \eqref{eq:median} is the optimal estimation in the maximum likelihood sense so that the observed data (objects) $O$ is most probable.
\corrected[2.7]{Although this elaboration is mathematically simple and known in some specific spaces (e.g. in the case of signed and unsigned permutations the generalized median is the maximum likelihood solution under the Mallows model \cite{tang_mallows_2019,arora_consensus_2013}), adapting it for the general case of arbitrary domains provides us a deeper understanding of the computation and we are not aware of its mention for general spaces in the literature before}.

Occasionally, researchers use the definition
\begin{align*}
\bar{o} = \arg \min_{o \in \mathcal{O}} \sum_{o_i \in O} \delta^2(o_i, o)
\end{align*}
in analogy of $(x-\bar{x})^2$ in the Gaussian distribution for $\mathbb{R}$ \cite{Hinarejos2001}. This can be uniformly treated using the definition \eqref{eq:median} by defining an auxiliary distance function $\delta^*() = \delta^2()$.

%%%%%%%%%%%%%%%%%%%%%%%%%%%%%%%%%%%%%%%%%%%%%%%%%%%%%%%%%%%%%%%%%%%%%%%%%%%%%
\section{\restructured{Framework of Embedding-Based Generalized Median Computation}}
\label{sec:history}

The embedding framework \cite{ferrer2010generalized,nienkoetter2019dpe} for computing the generalized median \eqref{eq:median} uses an explicit embedding function $\kernel(o_i)$ to transform the objects $o_i$ into vectors in $\mathbb{R}^d$, leading to the problem
\corrected[2.8]{%
\begin{align*}
\bar{x} = \arg \min_{x \in \mathbb{R}^d} \sum_{i=1}^{n} \lVert \kernel(o_i) - x \rVert_2
\end{align*}
}
in vector space using the Euclidean distance. This is solved by using the popular Weiszfeld algorithm \cite{Weiszfeld_2009}. This iterative algorithm delivers a good-quality geometric median in reasonable time. Then, an inverse transformation is used to find an object $\bar{o}$ with $\bar{o} \approx \kernel^{-1}(\bar{x})$ that can be seen as the inverse of $\kernel$, therefore solving
\begin{align*}
\bar{o} \approx \kernel^{-1}(\bar{x}) = \arg \min_{o \in \mathcal{O}} \lVert \kernel(o)-\bar{x} \rVert_2
\end{align*}
\corrected[2.9]{This is a special instance of the so-called pre-image problem in kernel-based machine learning \cite{Honeine11}} \correctedfinal[2]{which deals with the difficult reverse problem of finding an object in the original space corresponding to a given embedding.}

\restructured{In the following subsections we will first describe the embedding and median computation part of this framework resulting in $\bar{x}$, followed by an extensive review of the reconstruction step realizing the inverse transformation $\kernel^{-1}(\bar{x})$.}

\subsection{\restructured{Embedding and computation of generalized median in vector space}}

Prototype-based embedding was initially proposed in \cite{ferrer2010generalized}. Choosing $d$ prototype objects $p_1,...,p_d$ from the input set $O$, this method uses the embedding function
$$ \kernel(o_i) = (\delta(o_i,p_1), \ \delta(o_i,p_2), \ ..., \ \delta(o_i, p_d))$$
to compute vectors for each object. 

Later, it was shown that although easy to compute, the prototype embedding has several drawbacks \cite{nienkoetter2019dpe}, in particular the lacking preservation of distances in vector space. In that work, these drawbacks were remedied by using better distance-preserving embedding methods. Multidimensional scaling or curvilinear component analysis, for example, actively improve the preservation of pairwise distances between objects, thus leading to less approximation in the later steps and a better quality approximation of the median overall.
Although still only an approximation of the true distances, it was shown that using these and other embedding methods greatly improves the median quality through their accurate representation of the spatial structure of objects in vector space.

\restructured{Using the computed $\kernel(o_i)$ of each object, the generalized median $\bar{x}$ in vector space is then computed using the popular iterative Weiszfeld algorithm \cite{Weiszfeld_2009}
\begin{align}
\bar{x}_{j+1} =&  \frac{\sum_{i=1}^n \omega_i^j \kernel(o_i)}{\sum_{i=1}^n \omega_i^j} \nonumber
\intertext{with weights}
\omega_i^j =& \frac{1}{\lVert\bar{x}_j - \kernel(o_i)\rVert_2} \nonumber 
\end{align}
\corrected[2.12]{By selecting the starting vector $\bar{x}_0 \in \mathbb{R}^d$ in specific ways as shown in \cite{beck_weiszfelds_2015}, the Weiszfeld algorithm is proven to converge at a sublinear rate. In practice, however, it is often sufficient to use the mean of $\kernel(o_i)$ as easy-to-compute starting point since the algorithm quickly converges to a good approximation of $\bar{x}$ for all but a few trivial-to-resolve starting vectors \cite{kuhn1973note}.}}

\subsection{\restructured{Reconstruction of generalized median from vector space}}
\label{sec:reconstruction}

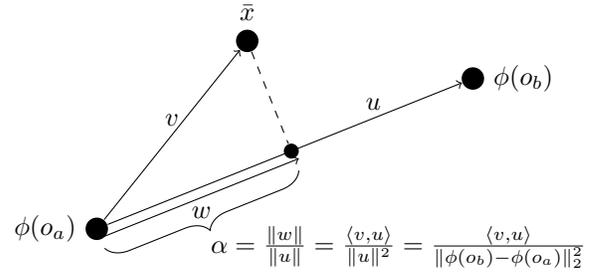
\begin{figure}
	\centering
	\begin{tikzpicture}

	% points
		\node[circle,fill,inner sep=3pt,label=left:$\phi(o_a)$] at (0,0) (a) {};
		\node[circle,fill,inner sep=3pt,label=right:$\phi(o_b)$] at (5,2) (b) {};
		\node[circle,fill,inner sep=3pt,label=above:$\bar{x}$] at (2,2.5) (m) {};

		% median
		\node[circle,fill,inner sep=2pt] at ($(a)!(m)!(b)$) (amb) {};
		
		% projection
%		\draw[dashed] (a) -- (b);
		\draw[dashed] (m) -- (amb);

		\draw[->] (a) -- node[above,pos=0.75] {$u$} (b);
		\draw[->] (a) -- node[above] {$v$} (m);

		\draw[->] ($(a)+(0.1,-0.1)$) -- node[below] {$w$} ($(amb)+(0.1,-0.1)$);
	
		\draw[decorate,decoration={brace, amplitude=3mm,mirror}] ($(a)!0.1!-90:(amb)$) -- node[midway,right,xshift=0mm,yshift=-5mm] {$\alpha = \frac{\lVert w \rVert}{\lVert u \rVert} = \frac{\inner{v}{u}}{\lVert u \rVert^2} = \frac{\inner{v}{u}}{\lVert\kernel(o_b) - \kernel(o_a)\rVert_2^2}$} ($(amb)!0.1!90:(a)$);
	
%	($(1)!(med)!(3)$) -- (1) node[midway,right,xshift=2mm,yshift=1mm] {$\alpha$};
\end{tikzpicture}
	\caption{Computation of the ratio $\alpha$ between objects for reconstruction.}
	\label{fig:projection}
\end{figure}

\restructured{The inverse function $\kernel^{-1}(\bar{x})$ is realized by a reconstruction process involving the nearest neighbors of $\bar{x}$ in the vector space.
The simplest reconstruction method -- called linear reconstruction -- chooses the two nearest neighbors of the median vector, and combines them in a similar fashion to linear interpolation, as seen in Figure \ref{fig:projection}. Let $\kernel(o_a), \kernel(o_b)$ be the nearest neighbors of median vector $\bar{x}$, $u = \kernel(o_b) - \kernel(o_a)$ and $v = \bar{x} - \kernel(o_a)$. Then, the reconstruction is achieved by projecting vector $v$ onto $u$, resulting in a vector $w = \frac{\inner{v}{u}}{\lVert u \rVert}$.
The resulting ratio
\[ \alpha = \frac{\lVert w \rVert}{\lVert u \rVert} = \frac{\inner{v}{u}}{\lVert \kernel(o_b) - \kernel(o_a)\rVert_2^2} \]
between $w$ and $u$ is then used to compute an approximate median object $o_m$ in the original space as an interpolation of $o_a, o_b$ with ratio $\alpha$. This is achieved with a so-called weighted mean function that constructs a $o_m$ using the following properties
\begin{align}
\delta(o_a,o_m) &= \alpha \cdot \delta(o_a,o_b) \nonumber\\
\delta(o_m,o_b) &= (1-\alpha) \cdot \delta(o_a,o_b) \label{eq:weighted-mean}
\end{align}
Under the assumption that the distances were reasonably well preserved in vector space, this approximation should have a lower SOD than the individual objects, similar to the case in vector space.
In practice, this weighted mean can often be derived from the distance function $\delta()$ \cite{Bunke_2001,Bunke_2002,Franek_2014}.}

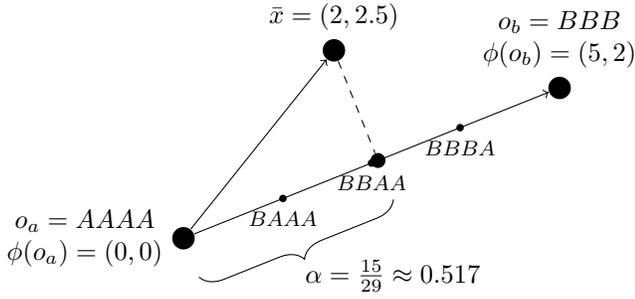
\begin{figure}
	\centering
	\begin{tikzpicture}

	% points
		\node[circle,fill,inner sep=3pt,label=left:\shortstack{$o_a=AAAA$\\$\phi(o_a)=(0,0)$}] at (0,0) (a) {};
		\node[circle,fill,inner sep=3pt,label=above:\shortstack{$o_b=BBB$\\$\phi(o_b)=(5,2)$}] at (5,2) (b) {};
		\node[circle,fill,inner sep=3pt,label=above:\shortstack{$\bar{x}=(2,2.5)$}] at (2,2.5) (m) {};

		% median
		\node[circle,fill,inner sep=2pt] at ($(a)!(m)!(b)$) (amb) {};
		
		% projection
%		\draw[dashed] (a) -- (b);
		\draw[dashed] (m) -- (amb);

		\draw[->] (a) -- node [circle, fill, inner sep=1pt, pos=0.25,label=below:{\footnotesize $BAAA$}] {} 
		                 node [circle, fill, inner sep=1pt, pos=0.5,label=below:{\footnotesize $BBAA$}] {} 
		                 node [circle, fill, inner sep=1pt, pos=0.75,label=below:{\footnotesize $BBBA$}] {}
		                 (b);
		\draw[->] (a) -- (m);

		%\draw[->] ($(a)+(0.1,-0.1)$) -- node[below] {$w$} ($(amb)+(0.1,-0.1)$);
	
		\draw[decorate,decoration={brace, amplitude=3mm,mirror}] ($(a)!0.2!-90:(amb)$) -- node[midway,right,xshift=0mm,yshift=-5mm] {$\alpha = \frac{15}{29} \approx 0.517$} ($(amb)!0.2!90:(a)$);
	
%	($(1)!(med)!(3)$) -- (1) node[midway,right,xshift=2mm,yshift=1mm] {$\alpha$};
\end{tikzpicture}
	\caption{\corrected[2.1]{}\corrected[2.10]{}\corrected[2.13]{Example for the weighted mean computation using strings. The median object $\hat{o}$ is reconstructed from $\bar{x}$ using the two neighbors $o_a = AAAA$ and $o_b = BBB$ and their respective embeddings. For visualization, possible reconstructed objects using a weighted mean function are shown as points on the line between $\kernel(o_a)$ and $\kernel(o_b)$.}}
	\label{fig:wm-example}
\end{figure}

\corrected[2.1]{}\corrected[2.9]{}\corrected[2.10]{}\corrected[2.13]{An example using strings and the common Levenshtein edit distance can be seen in Figure \ref{fig:wm-example}. Given a string set $O$, an embedding $\kernel(o)$ was computed that assigns each string a vector in a two-dimensional space. In this case, object $o_a = AAAA$ was assigned $\kernel(o_a)=(0,0)$ and object $o_b=BBB$ was assigned $\kernel(o_b)=(5,2)$. Similarly, all other objects in the set were assigned embedded vectors. After computation of the generalized median in vector space $\bar{x} = (2,2.5)$, $\kernel(o_a)$ and $\kernel(o_b)$ were identified as nearest objects. Then, the linear reconstruction method projects $\bar{x}$ onto the line between $\kernel(o_a)$ and $\kernel(o_b)$ to compute a ratio $\alpha \approx 0.517$. The unknown pre-image $\kernel^{-1}(\bar{x})$ is then approximated by constructing a new object $o_m$ that approximates the properties shown in Eq. (\ref{eq:weighted-mean}). In the case of the Levenshtein edit distance, a list of minimal necessary edit operations can be obtained to transform $o_a$ into $o_b$, in this case for example $AAAA \rightarrow BAAA \rightarrow BBAA \rightarrow BBBA \rightarrow BBB$ with corresponding edit distance 4 assuming that all editing steps have cost 1. Therefore, $BBAA$ is selected as approximation of $\kernel^{-1}(\bar{x})$ since it is the closest of the intermediate edit steps to fulfilling Eq. (\ref{eq:weighted-mean}) with $\delta(o_a,o_m) = 2 \approx 2.06 = \alpha \cdot \delta(o_a,o_b)$ and $\delta(o_m,o_b) = 2 \approx 1.93 = (1-\alpha) \cdot \delta(o_a,o_b)$. This can be understood as individual editing steps lying on the line between $\kernel(o_a)$ and $\kernel(o_b)$ and choosing the closest one to the projected $\bar{x}$.}

This type of reconstruction relies heavily on the fact that in vector space, the projection of a vector $\bar{x}$ onto the line between $\kernel(o_a)$ and $\kernel(o_b)$ is the closest point on this line to $\bar{x}$ (see Figures \ref{fig:projection} \restructured{and \ref{fig:wm-example})}.
In embedding based frameworks this is assumed to be true in object space as well and approximated by the weighted mean. Just as in vector space where any point on a line can be expressed as a weighted mean between its end points, a "straight line" in object space can be represented by all objects fulfilling the weighted mean Equation (\ref{eq:weighted-mean}). However, depending on the distance formulation, this "line" may not be unique. \corrected[2.11]{Using the example above, a second "straight line" of editing steps is $AAAA \rightarrow AAAB \rightarrow AAB \rightarrow BAB \rightarrow BBB$, leading to the weighted mean $AAB$. This can even happen in vector space using certain distances, for example the Manhattan metric, where an infinite number of possible paths of weighted means exist between two vectors.}

\corrected[2.11]{Using this definition of "straight line", the orthogonal projection of an object onto the line between two objects is the (not necessarily unique) object which is a weighted mean between $o_a$ and $o_b$ and has the smallest distance to the projected object.
However, for many popular distance formulations computing this projection would require to compute a large number of projection candidates that could be very costly itself.} For more efficiency of these methods, often only a small number of candidates are computed and the best one is returned. In our case, we will balance the computation of the weighted mean between speed and robustness by only computing two weighted means. The weighted mean from $o_a$ to $o_b$ using $\alpha$ and the weighted mean from $o_b$ to $o_a$ using $1-\alpha$ are computed, choosing the weighted mean with the better approximation of the median as the result. This is done for both the explicit generalized median computation shown in this section and our kernel based method shown later for a fair comparison.

In addition to the simple linear reconstruction, there are other reconstruction methods, named triangular and recursive reconstruction. The basic idea of these methods is the same, but using three objects instead of two nearest neighbors for triangular reconstruction, or using a more sophisticated projection technique for recursive reconstruction (see \cite{ferrer2010generalized,nienkoetter2019dpe} for details).

Further improvement of the inverse transformation can be found in \cite{nienkoetter2016reconstruction}. Two new reconstruction methods were proposed, namely linear-recursive and triangular-recursive reconstruction, as well as a method that iteratively improves results of any reconstruction method. 
Both linear-recursive and triangular-recursive reconstruction still use the same weighted mean function and projection technique, but combine objects in a different way.
%linear-recursive reconstruction for example groups all objects in pairs and computes a weighted mean for each. This new set of objects is then recursively combined until only one object remains. The result of the combination is the object with the lowest sum of distances.
It could be shown that these methods significantly improve the quality of the reconstructed median over all three previous variants by enhancing the inverse transformation $\kernel^{-1}(\bar{x})$.

In summary, the framework of embedding-based generalized median computation comprises three steps: applying an explicit method $\kernel(o_i)$ to embed objects from an arbitrary domain into a vector space, computing the median $\bar{x}$ there, and then using an inverse transformation $\kernel^{-1}(\bar{x})$ to compute an approximate median object in the original space that corresponds to this median vector.

Despite the popularity and the recent advances, the prototype-embedding approach \cite{ferrer2010generalized,nienkoetter2019dpe} has some inherent drawbacks due to its nature of \textit{explicit} transformation for the embedding purpose. There is a need of specifying the embedding dimension (number of prototypes) and selecting this number of suitable prototypes. Both may influence the quality of the computed generalized median, but are not always trivial to set. Therefore, we aim to explore instead an approach of \textit{implicit} transformation by means of kernel functions. This approach helps overcome the inherent drawbacks of explicit transformation. More importantly, some of the studied kernel functions have the nice property of provably preserving the pairwise distances after embedding in the implicit vector space, which is fundamental for superior performance of generalized median computation within the embedding-based computation framework \cite{nienkoetter2019dpe}.

%%%%%%%%%%%%%%%%%%%%%%%%%%%%%%%%%%%%%%%%%%%%%%%%%%%%%%%%%%%%%%%%%%%%%%%%%%%%%
\section{Kernel-Based Generalized Median Computation}
\label{sec:kernel-mean}

%As shown in the previous section, the main idea of reconstructing median objects in an arbitrary domain is finding an accurate way to compute the spatial relationship between objects and their median, without knowing the median object.

In this section we present a kernel approach to embedding-based generalized median computation. Despite the {\it implicit} nature of kernel embedding functions it turns out to be possible to determine all required ingredients for this computation. In particular, we will show that the ratio $\alpha$ that is necessary for the reconstruction combination of objects in $\kernel^{-1}()$ can be computed without using an explicit embedding $\kernel()$, a kernel function suffices instead. Using the kernel approach, one thus can accurately reconstruct a generalized median without the need of an explicit vector space.

After a brief introduction to kernel methods in Section \ref{subsec:kernel-introduction}, we shortly repeat the main steps of the generalized median computation using explicit transformation in Section \ref{subsec:scheme}. The main contribution of this work is then described in Section \ref{subsec:gen-med-kernel} (positive definite kernels) and Section \ref{subsec:gen-med-indefinite-kernel} (indefinite kernels) to perform generalized median computation using implicit transformation in the context of kernel functions. Then, we present several kernel functions in Section \ref{subsec:kernel-functions} that will be studied in our experimental work. \restructured{An analysis of the complexity of the kernel-based median computation framework is given in Section \ref{subsec:complexity}}. Finally, we provide some further discussion in Section \ref{subsec:notes-approximation}.

\subsection{Kernel methods}
\label{subsec:kernel-introduction}

Kernel functions are well-known from their application in kernel machine \cite{Ma2019}, support vector machine (SVM), clustering, principal component analysis, etc. \cite{Kernel_Book_2004}. Originally, a kernel function $K(x,y)$ is defined by
\[ K\!: \ \mathbb{R}^d\times\mathbb{R}^d \rightarrow \mathbb{R} \]
with a related transformation $\kernel\!: \mathbb{R}^d \rightarrow \mathcal{H}_K$ into a Hilbert space $\mathcal{H}_K$ so that $K(x,y) = \inner{\kernel(x)}{\kernel(y)}$, that is, the kernel function corresponds to the scalar product in $\mathcal{H}_K$. One simple example for the two-dimensional vector space is the kernel function
\[ K\!: \ \mathbb{R}^2\times\mathbb{R}^2 \rightarrow \mathbb{R}, ~K(x,y) = (x y^\top)^2 \]
which computes the scalar product
\[ \inner{\kernel(x)}{\kernel(y)} = \inner{(x_1^2, \sqrt{2} x_1 x_2, x_2^2)}{(y_1^2, \sqrt{2} y_1 y_2, y_2^2)}\]
of a transformation in a three-dimensional vector space. In general, using a kernel function $K(x,y)$, one can compute the scalar product in the transformed vector space without the need of explicitly transforming $x$ and $y$ into $\kernel(x)$ and $\kernel(y)$, respectively. Since the linear SVM classifier depends on scalar products only, this \textit{kernel trick} allows nonlinear classification of the original vectors by a linear classification in the vector space implied by a kernel function.

The concept of kernel functions for vector space can be extended to an arbitrary space $\mathcal{O}$ to enable 
classification on other types of data (e.g. strings, trees, graphs) \cite{gartner2003survey,Gaertner_book,Xu18,Ghosh18}. In this general case, a kernel function is defined by
\begin{align*}
K\!: \ \mathcal{O}\times\mathcal{O} \rightarrow \mathbb{R}, ~K(o_a,o_b) = \inner{\kernel(o_a)}{\kernel(o_b)}
\end{align*}
with a transformation $\kernel\!: \mathcal{O} \rightarrow \mathcal{H}_K$ that computes the scalar product of the transformed vectors of two objects $o_a$ and $o_b$. Note that many of these kernels do not have an explicit representation of a vector space and compute the scalar product directly instead.

Since the scalar product in a Hilbert space induces a norm, the distance between two transformed vectors $\kernel(o_a)$ and $\kernel(o_b)$ can be computed using the kernel function only as follows
\begin{align}
\lVert \kernel(o_a) - \kernel(o_b)\rVert_2 =& \left(\inner{\kernel(o_a)-\kernel(o_b)}{\kernel(o_a)-\kernel(o_b)}\right)^\frac{1}{2} \nonumber \\
=& \left(\inner{\kernel(o_a)}{\kernel(o_a)}-2\inner{\kernel(o_a)}{\kernel(o_b)} \right. \nonumber\\
&\left.+\inner{\kernel(o_b)}{\kernel(o_b)}\right)^{\frac{1}{2}} \nonumber \\
=& \left(K(o_a,o_a) - 2 K(o_a,o_b) + K(o_b,o_b)\right)^\frac{1}{2}
\label{eq:kernel-norm}
\end{align}

\subsection{Generalized median computation using explicit transformation}
\label{subsec:scheme}

\restructured{To better understand the kernel-based reconstruction method, we will shortly repeat the central elements of the explicit transformation framework as presented in Section \ref{sec:history}.}
In case of explicitly known transformation $\kernel$ as with the previous prototype and distance-preserving embedding methods \cite{nienkoetter2019dpe}, the Weiszfeld algorithm \cite{Weiszfeld_2009} is applied to iteratively compute the generalized median $\bar{x}$ in vector space by
\begin{align}
\bar{x}_{j+1} =&  \frac{\sum_{i=1}^n \omega_i^j \kernel(o_i)}
{\sum_{i=1}^n \omega_i^j} \label{eq:weiszfeld_weights1}
\intertext{with weights}
\omega_i^j =& \frac{1}{\lVert\bar{x}_j - \kernel(o_i)\rVert_2} \label{eq:weiszfeld_weights2}
\end{align}
\restructured{using explicit vectors $\kernel(o_i)$ for each object $o_i$ in the input set and nearly any starting vector $\bar{x}_0$.
Then, a ratio $\alpha$ of the projection of $\bar{x}$ onto the line between its closest objects $\kernel(o_a)$ and $\kernel(o_b)$ is computed by}
\begin{equation}
\alpha  %= \frac{\lVert w \lVert_2}{\lVert u \lVert_2}
           %= \frac{\inner{v}{u}}{\lVert u \rVert^2}
	        = \frac{\inner{\bar{x}-\kernel(o_a)}{\kernel(o_b)-\kernel(o_a)}}{\lVert\kernel(o_b)-\kernel(o_a)\rVert_2^2}
\label{eq:projection_alpha}
\end{equation}
The generalized median $\bar{o}$ is finally reconstructed by computing a weighted mean object between $o_a$ and $o_b$ using this ratio \restructured{as shown in Section \ref{sec:reconstruction}}.

\subsection{Generalized median computation using implicit transformation: positive definite kernels}
\label{subsec:gen-med-kernel}

The computation scheme above explicitly needs the transformation $\kernel(o_i)$ for each object $o_i$ in the input set and the intermediate result $\bar{x}_j$ after each iteration of the Weiszfeld algorithm. When working with positive definite kernel functions along with an implicit transformation $\kernel$, however, these fundamental ingredients for median computation are not available. In the following we present a solution to this problem so that we can still use the scheme presented in Section \ref{subsec:scheme} to compute the generalized median.

\vspace{2mm}
\noindent
{\bf Median computation in kernel space:} 
Let $\bar{x}_j$ be the generalized median vector from the Weiszfeld algorithm at iteration $j$, see (Eq. (\ref{eq:weiszfeld_weights1})). Then, the weight $\omega_i^j$ can be computed using Eq. (\ref{eq:weiszfeld_weights2}) and (\ref{eq:kernel-norm}) as
\begin{equation}
\omega_i^j = 1 / (\inner{\bar{x}_j}{\bar{x}_j} - 2 \inner{\bar{x}_j}{\kernel(o_i)} + \inner{\kernel(o_i)}{\kernel(o_i)})^\frac{1}{2}
\label{eq:kernel-weiszfeld1}
\end{equation}
Here $\inner{\kernel(o_i)}{\kernel(o_i)}$ can be computed using the kernel function $K(o_i,o_i)$. On the other hand, since $\bar{x}_j$ is unknown we are unable to evaluate $\inner{\bar{x}_j}{\bar{x}_j}$ and $\inner{\bar{x}_j}{\kernel(o_i)}$. Nevertheless, these two items can be precisely determined using the iterative nature of the Weiszfeld algorithm by substituting $\bar{x}_j$ with its previous step of Eq. (\ref{eq:weiszfeld_weights1}) and using the bilinear properties of the scalar product. This results in
\begin{align}
\inner{\bar{x}_j}{\bar{x}_j} =&\inner{\frac{\sum_{u=1}^n \omega_u^{j-1} \kernel(o_u)}{\sum_{u=1}^n \omega_u^{j-1}}}{\frac{\sum_{u=1}^n \omega_u^{j-1} \kernel(o_u)}{\sum_{u=1}^n \omega_u^{j-1}}} \nonumber \\
%=& \left. \left( \sum_{k=1}^{n} \sum_{l=1}^{n} \omega_k^{j-1} \omega_l^{j-1} K(o_k,o_l) \right) \right. \nonumber \\ 
%& \left.\middle/\left( \sum_{k=1}^{n} \omega_k^{j-1} \right)^2\right. \label{eq:kernel-weiszfeld3}
=& \frac{\sum_{u=1}^{n} \sum_{v=1}^{n} \omega_u^{j-1} \omega_v^{j-1} K(o_u,o_v)}{\left(\sum_{u=1}^{n} \omega_u^{j-1} \right)^2} \label{eq:kernel-weiszfeld2}
\end{align}
and
\begin{align}
\inner{\bar{x}_j} {\kernel(o_i)} =& \inner{\frac{\sum_{u=1}^n \omega_u^{j-1} \kernel(o_u)}{\sum_{u=1}^n \omega_u^{j-1}}}{\kernel(o_i)} \nonumber \\
=& \frac{\sum_{u=1}^{n} \omega_u^{j-1} K(o_u,o_i)}{\sum_{u=1}^{n} \omega_u^{j-1}}
\label{eq:kernel-weiszfeld3}
\end{align}
Note that the right hand side of Eq. (\ref{eq:kernel-weiszfeld2}) and (\ref{eq:kernel-weiszfeld3}) only contains kernels between input objects and the weights of the Weiszfeld algorithm in the previous step of the iteration. 
Bringing Eq. (\ref{eq:kernel-weiszfeld1}), (\ref{eq:kernel-weiszfeld2}) and (\ref{eq:kernel-weiszfeld3}) together, one can therefore compute  $\omega_i^j$ iteratively from kernels between input objects without explicit knowledge of the transformed vectors $\kernel(o_i)$ and the immediate median $\bar{x}_j$. As a starting value for this iteration scheme, one can for example set all $\omega_i^0 = 1$. This corresponds to the mean of objects in kernel space and is often a good first approximation of the median result. It is important to emphasize that we are only able to use the Weiszfeld algorithm to compute the weights $\omega_i^j$, but not the immediate median $\bar{x}_j$. As will be shown in the following, this is sufficient for reconstructing the generalized median in the original space.

\vspace{2mm}
\noindent
{\bf Reconstruction:}
Even without explicitly knowing the median $\bar{x}$ in the kernel space, we need to reconstruct the generalized median $\bar{o} = \kernel^{-1}(\bar{x})$ in the original space. Our solution starts with the insight that for the reconstruction, we actually do not really need to explicitly know the median $\bar{x}$. It suffices to know the nearest neighbors of $\bar{x}$ and the related ratio $\alpha$ in Eq. (\ref{eq:projection_alpha}). Their corresponding input objects are then combined to build the generalized median $\bar{o}$. The nearest neighbors of $\bar{x}$ can be easily found by the final weights $\omega_i$ after convergence of the Weiszfeld algorithm since Eq. (\ref{eq:weiszfeld_weights2}) implies that these weights are the inverse of the norm between object $o_i$ and the median in kernel space. That is, sorting the final weights $\omega_i$ in a descending order will deliver the nearest neighbors of the unknown $\bar{x}$.

%Explicit kernel \cite{Vega_PR10}. String (i.e. strings with known lengths \cite{Bakir03}, $n$-gram based generic string kernel \cite{GiguerePlos15} and its special cases with additional constraints of one of the control parameters equal 0 \cite{GiguereICML15}). Graphs (the marginalized graph kernel) \cite{Bakir04}
%}

The final weights $\omega_i$ are related to the unknown median $\bar{x}$ in kernel space by
\begin{equation}
\bar{x} =  \frac{\sum_{i=1}^n \omega_i \kernel(o_i)} {\sum_{i=1}^n \omega_i}
\label{eq:convergence}
\end{equation}
Given two nearest neighbors $o_a$ and $o_b$, inserting Eq. (\ref{eq:convergence}) into Eq. (\ref{eq:projection_alpha}) leads to
\begin{align}
\alpha =& \frac{\inner{\frac{\sum_{i=1}^n \omega_i \kernel(o_i)}{\sum_{i=1}^{n} \omega_i}-\kernel(o_a)}{\kernel(o_b)-\kernel(o_a)}}{||\kernel(o_b)-\kernel(o_a)||^2}  \nonumber\\
=& \frac{\frac{\sum_{i=1}^{n} \omega_i \left(K(o_i,o_b) - K(o_i,o_a)\right)}{\sum_{i=1}^{n} \omega_i} - K(o_a,o_b) + K(o_a,o_a)}{K(o_b,o_b)-2 K(o_b,o_a) + K(o_a,o_a)}  \nonumber \\ \label{eq:kernel-alpha}
\end{align}
This computation only needs the final weights $w_i$ and kernel values between all objects in the set as well as $o_a$ and $o_b$.

The simplest method, linear reconstruction \cite{nienkoetter2019dpe,ferrer2010generalized}, uses the two nearest neighbors $o_1$ and $o_2$ for the reconstruction. Starting with the -- in kernel space -- closest object to the median $\bar{o}_1 = o_1$, the $\alpha$ value between it and the next closest object $o_2$ is computed and applied in the weighted mean function to generate a better approximation $\bar{o}_2$ of the median object. For the triangular reconstruction, a third nearest neighbor $o_3$ is needed and another $\alpha$ is computed between the median object $\bar{o}_2$ and $o_3$ towards a refined median object $\bar{o}_3$. The returned median approximation is the last constructed object $\bar{o}_l$. This is shown in Algorithm \ref{alg:kernel-linear}, using $l=2$ for linear and $l=3$ for triangular reconstruction.

\begin{algorithm}[t]
	\caption{Kernel-based Linear ($l=2$) and Triangular ($l=3$) Reconstruction}
	\label{alg:kernel-linear}
	
	\begin{algorithmic}[1]
		\renewcommand{\algorithmicrequire}{\textbf{Input:}}
		\renewcommand{\algorithmicensure}{\textbf{Output:}}
		
		\REQUIRE Object set $O$, integer $l$, final weights $\omega_i$, weighted mean function $wm()$
		\ENSURE Median object $\bar{o}$
		
		\STATE Select $o_1, \dots, o_l$ with the $l$ maximal $\omega_i$
		\STATE $\bar{o}_1 = o_1$
		\FOR {$j=2$ to $l$}
		\STATE Compute $\alpha$ using Eq. (\ref{eq:kernel-alpha}) with objects $\bar{o}_{j-1}$, $o_j$
		\STATE $\bar{o}_j = wm(\bar{o}_{j-1}, o_j, \alpha)$
		\ENDFOR
		\RETURN $\bar{o}_l$ with the related SOD
	\end{algorithmic}
\end{algorithm}

\begin{algorithm}
	\caption{Kernel-based Linear (Triangular) Recursive Reconstruction}
	\label{alg:kernel-lin-rec}
	
	\begin{algorithmic}[1]
		\renewcommand{\algorithmicrequire}{\textbf{Input:}}
		\renewcommand{\algorithmicensure}{\textbf{Output:}}
		
		\REQUIRE Object set $O$, final weights $\omega_i$, weighted mean function $wm()$
		\ENSURE Median object $\bar{o}$
		
		\STATE $\bar{o}_{best} = \emptyset$
		\WHILE{$|O| > 1$}
		\STATE Divide $O$ into $|O|/2$ pairs ($|O|/3$ triples) by grouping maximal $\omega_i$ first.
		\STATE $O' = \emptyset$
		\FOR {each pair $(o_a,o_b)$ (triple $(o_a,o_b,o_c)$)}
		\STATE Compute $\bar{o}$ using Algorithm \ref{alg:kernel-linear} with $l = 2$ ($l = 3$)
		\STATE $O' = O' \cup \{\bar{o}\}$
		\ENDFOR
		\STATE $\bar{o}_{best} = \arg \displaystyle\min_{o \in \{\bar{o}_{best}\} \cup O'} SOD(o)$
		\STATE $O = O'$ 
		\ENDWHILE
		
		\RETURN $\bar{o}_{best}$ with the related SOD
	\end{algorithmic}
\end{algorithm}

We were not able to adapt the recursive and best-recursive reconstruction methods \cite{ferrer2010generalized,nienkoetter2019dpe} for kernel methods due to the recursive projection onto hyperplanes. Therefore, we leave these methods out and instead adapted linear recursive and triangular recursive reconstruction described in \cite{nienkoetter2016reconstruction}. For prototype embedding, they show superior results compared to best-recursive reconstruction anyway and therefore should have a similar performance for kernel methods. Both methods are shown in Algorithm \ref{alg:kernel-lin-rec}. First, the object set is divided into pairs (or triples) of objects, and a linear (triangular) reconstruction is performed for each case. This results in a new set of $|O|/2$ ($|O|/3$) objects. The algorithm is repeated using this new set until only one object remains. The object with the lowest SOD of all computed objects is returned as the approximated median.

%To summarize, we showed in this section that one can compute the ratio $\alpha$ of the approximated generalized median for any two objects $o_a$ and $o_b$ in vector space solely using kernel functions, without the need to compute an explicit embedding or median vector. In the next section, we will show how to use this in various strategies to compute an approximate generalized median from neighboring objects.

\vspace{2mm}
\noindent
{\bf Overall algorithm:}
The overview of the proposed kernel-based generalized median framework is shown in Algorithm \ref{alg:kernel-median}. First in lines 1 to 6, the necessary weights of the objects are determined using one of the kernel methods to be discussed in Section \ref{subsec:kernel-functions}. Note that in contrast to previous explicit embedding, neither the vectors $\kernel(o_i)$ nor the median vector $\bar{x}$ can be explicitly computed. As for the number $j_{max}$ of iterations of the kernel-Weiszfeld algorithm, only a low number of iterations is needed in practice and the computation can be stopped early once $\omega_i^j$ converges (see Section \ref{subsec:evaluation-convergence}). These weights are then used in one of the reconstruction methods to compute an approximate generalized median in the original space.

\begin{algorithm}
	\caption{Kernel-Based Generalized Median Framework}
	\label{alg:kernel-median}
	
	\begin{algorithmic}[1]
		\renewcommand{\algorithmicrequire}{\textbf{Input:}}
		\renewcommand{\algorithmicensure}{\textbf{Output:}}
		
		\REQUIRE Object set $O$, distance function $\delta()$, weighted mean function $wm()$, kernel function $K()$
		\ENSURE Median object $\bar{o}$
		\\ /* \textit{Computation of median weights} */
		\STATE Initialize $\omega_i^0 = 1$ for all $1 \leq i \leq |O|$
		\FOR {$j = 1$ to $j_{max}$}
			\FOR {$i = 1$ to $n$}
				\STATE Compute $\omega_i^j$ using Eq. (\ref{eq:kernel-weiszfeld1}), (\ref{eq:kernel-weiszfeld2}) and (\ref{eq:kernel-weiszfeld3})
			\ENDFOR
		\ENDFOR
		\\ /* \textit{Reconstruction} */
		\STATE Compute $\bar{o}$ using reconstruction algorithm \ref{alg:kernel-linear} or \ref{alg:kernel-lin-rec}
		\RETURN $\bar{o}$ with the related SOD
	\end{algorithmic}
\end{algorithm}

\subsection{Generalized median computation using implicit transformation: indefinite kernels}
\label{subsec:gen-med-indefinite-kernel}

In the previous section we have shown the computation of the generalized median in kernel space using positive definite kernel functions. However, a number of kernel functions that can be applied to any domain given a distance function (as shown in Section \ref{subsec:kernel-functions} and later used in our evaluation) are not positive definite and therefore do not fulfill some basic assumptions used in the previous section.  This is similar to the case of kernel SVM, where these kernels are used for classification even though they are not guaranteed to be positive definite \cite{haasdonk2005feature}. To ensure the general nature of this framework, we will attempt to approximate the generalized median using these indefinite kernels as well. For this reason we will shortly introduce the concept of pseudo-Euclidean spaces and show that indefinite kernel functions can be used to compute the generalized median in such a pseudo-Euclidean space instead of normal Euclidean space.

Pseudo-Euclidean spaces are linear vector spaces including an indefinite, symmetric bilinear form $\inner{\cdot}{\cdot}$ and can be expressed as $\mathbb{E} = \mathbb{R}^{(p,q)} = \mathbb{R}^p \times i\mathbb{R}^q$, where $i$ is the imaginary unit, i.e. as spaces whose vectors consist of $p$ real and $q$ imaginary elements \cite{pekalska2001generalized}. As such, pseudo-Euclidean spaces are subspaces of the complex vector space $\mathbb{C}^{p+q}$. In contrast to standard Euclidean spaces, this allows for negative squared distances
$\delta^2: O \times O \rightarrow \mathbb{R}$ with
\begin{align}
    \delta^2(o_i,o_j) &=  \inner{x_i}{x_i} - 2 \inner{x_i}{x_j} + \inner{x_j}{x_j}.
    \nonumber%\label{eq:indefinite-square-distance}
\end{align}

According to \cite{haasdonk2005feature,pekalska2001generalized}, any symmetric indefinite kernel function $K(o_i,o_j)$ can be used to construct such a real squared distance by substituting the inner product with indefinite kernels
\begin{align}
    \delta^2(o_i,o_j) = K(o_i,o_i) - 2 K(o_i,o_j) + K(o_j,o_j) 
    \label{eq:indefinite-square-distance-kernel}
\end{align}
Additionally, if this squared distance is symmetric with $\delta^2(o_i,o_i) = 0$ for all $x \in O$, which is true for Equation (\ref{eq:indefinite-square-distance-kernel}) for any symmetric function $K$, one can express $\delta^2$  as
\begin{align}
    \delta^2(o_i,o_j) = \lVert \kernel(o_i) - \kernel(o_j) \rVert_2^2
    \label{eq:indefinite-square-norm}
\end{align}
using a transformation $\kernel : O \rightarrow \mathbb{E}$ into a pseudo-Euclidean space $\mathbb{E}$.
As such, indefinite kernels allow the computation of a real-valued squared norm in a pseudo-Euclidean space, which, as the square root of a real number, translates to a complex valued norm consisting only of a real \textit{or} imaginary part.

As we need the norm in the computation of the weights of the Weiszfeld algorithm as shown in Equation (\ref{eq:weiszfeld_weights2}), we will need to use the root of the above squared distance shown in Equation (\ref{eq:indefinite-square-norm}) and as such expand the pseudo-Euclidean space to a full complex space. This leads to modifications in Equations (\ref{eq:kernel-weiszfeld1}), (\ref{eq:kernel-weiszfeld2}) and (\ref{eq:kernel-weiszfeld3}) for the iterative computation of weights using complex norms. In contrast to real spaces, the inner product in complex spaces is conjugate symmetric ($\inner{x}{y} = \overline{\inner{y}{x}}$) and conjugate linear in the second argument ($\inner{x}{\lambda y} = \overline{\lambda} \inner{x}{y}$).

Using the conjugate symmetry, Equation (\ref{eq:kernel-weiszfeld1}) becomes
\begin{align}
\omega_i^j =& 1 / (\inner{\bar{x}_j}{\bar{x}_j} - \inner{\bar{x}_j}{\kernel(o_i)} \nonumber \\
&- \overline{\inner{\bar{x}_j}{\kernel(o_i)}} + \inner{\kernel(o_i)}{\kernel(o_i)})^\frac{1}{2}
\label{eq:kernel-weiszfeld1-complex}
\end{align}
while, using conjugate linearity in the second argument, Equation \ref{eq:kernel-weiszfeld2} becomes
\begin{align}
\inner{\bar{x}_j}{\bar{x}_j} =&\inner{\frac{\sum_{u=1}^n \omega_u^{j-1} \kernel(o_u)}{\sum_{u=1}^n \omega_u^{j-1}}}{\frac{\sum_{u=1}^n \omega_u^{j-1} \kernel(o_u)}{\sum_{u=1}^n \omega_u^{j-1}}} \nonumber \\
%=& \left. \left( \sum_{k=1}^{n} \sum_{l=1}^{n} \omega_k^{j-1} \omega_l^{j-1} K(o_k,o_l) \right) \right. \nonumber \\ 
%& \left.\middle/\left( \sum_{k=1}^{n} \omega_k^{j-1} \right)^2\right. \label{eq:kernel-weiszfeld3}
=& \frac{\sum_{u=1}^{n} \sum_{v=1}^{n} \omega_u^{j-1} \overline{\omega_v^{j-1}} K(o_u,o_v)}{\left(\sum_{u=1}^{n} \omega_u^{j-1}\right) \left(\overline{\sum_{u=1}^{n} \omega_u^{j-1}}\right) } \label{eq:kernel-weiszfeld2-complex}
\end{align}
Equation (\ref{eq:kernel-weiszfeld3}) remains unchanged. Note that since the complex conjugate does not affect real numbers, these equations can also be used for positive definite kernels. As these modifications may influence the convergence of the Weiszfeld algorithm, we will experimentally study its convergence using indefinite kernels in Section \ref{subsec:evaluation-convergence}.

All in all, this leads to complex weights in the computation of the Weiszfeld algorithm, which -- inserted into the also unchanged Equation (\ref{eq:kernel-alpha}) -- can lead to a complex $\alpha$ value. However, for the computation of the weighted mean real value is needed. As such, in the case of a complex $\alpha$, we will use its magnitude to compute the weighted mean.

\subsection{Kernel functions}
\label{subsec:kernel-functions}

We briefly describe a variety of kernel functions used in our experimental work, both positive definite domain-dependent kernels and indefinite domain-independent kernels. In addition, we show that several special cases of these kernel functions are able to preserve the distance in the vector space of their respective projection, thus making them ideal candidates for kernel-based median computation.

For a general framework for generalized median computation one would like to use distance-preserving domain-independent positive definite kernel functions to ensure a good distance-preserving embedding in an implicit kernel space where the median is approximated, similar to the argument for distance-preserving explicit embedding methods in \cite{nienkoetter2019dpe}. However, we were not able to find appropriate positive definite kernel functions that could be used for every domain. Therefore, we will first present a number of positive definite domain-dependent kernels that can be used in this framework, followed by a number of indefinite kernel functions that can be used in any domain. In the end, we will discuss the distance preservation of these kernels.

\subsubsection{Positive definite domain-dependent kernels}
\label{subsec:kernel-functions:pos-definite}

Although positive definite domain-independent kernels would be preferable for our method shown in Section \ref{subsec:gen-med-kernel}, we were only able to find positive definite kernels for specific domains. Therefore, we will present three domain specific kernel functions in this section. 

\vspace{1mm}
\noindent
\textbf{String subsequence kernel:}
\corrected[1.4]{The string subsequence kernel $K^{ssk}$ \cite{lodhi2002text} is a popular kernel function in string based applications for strings $s$ of alphabet $\Sigma$. Given the set $\{u_1, u_2, ...\} \subset \Sigma^{|u|}$ of all possible substrings $u_i$ with fixed length $|u|$, the transformation of a string $s$ is defined as 
\begin{align*}
    \kernel^{ssk}(s) &= (\kernel_{u_1}(s), \kernel_{u_2}(s), ...), \quad
    \kernel_u(s) &= \sum_{\mathbf{i} : u = s[\mathbf{i}]} \lambda^{l(\mathbf{i})}
\end{align*}
where $\mathbf{i} =(i_1, ..., i_{|u|})$ are increasing indices, $s[\mathbf{i}]$ the substring of $s$ consisting only of the symbols at positions in $\mathbf{i}$, and $l(\textbf{i}) = i_{|u|} - i_1 + 1$ is the length of the substring containing $\mathbf{i}$. $\lambda \leq 1$ is a weighting parameter. As such, each value in vector $\kernel^{ssk}(s)$ encodes the frequency and compactness of the occurrences of a substring $u_i$ in string $s$. Interestingly, the kernel $K^{ssk}(s_1,s_2) = \inner{\kernel^{ssk}(s_1)}{\kernel^{ssk}(s_2)}$ can be computed recursively without the need to explicitly compute the embedding $\kernel^{ssk}(s)$ \cite{lodhi2002text}. In our evaluation, we chose $|u| = 2$ and $\lambda = 0.5$.}

\vspace{1mm}
\noindent
\textbf{Partition kernel:}
The partition kernel is designed for the domain of clusterings and based on a simple transformation
\begin{align*}
\kernel^{part}(x) &= (\kernel^{p}_{1,2}(x),...,\kernel^{p}_{1,n}(x),\kernel^{p}_{2,3}(x),...,\kernel^{p}_{n,n}(x)) \\
\kernel^{p}_{i,j}(x) &= \begin{cases}
1, & x(i) = x(j)\\
0, & \mbox{otherwise}
\end{cases}
\end{align*}
where $x$ is a vector containing the clustering labels of $n$ elements. If done for two clusterings, the scalar product $K^{part}(o_a,o_b) = \inner{\kernel^{part}(o_a)}{\kernel^{part}(o_b)}$ is the number of pairs whose cluster labels are equal in $o_a$ and $o_b$, while the label itself can be different. As a scalar product between two vector transformations, it is guaranteed to be a positive definite kernel function.

\vspace{1mm}
\noindent
\textbf{Kendall kernel:}
The Kendall kernel $K^{kend}$ \cite{jiao2017kendall} is designed for the domain of permutations and based on the Kendall-$\tau$ distance between permutations. This distance counts the number of pairs whose order is different between two permutations. For two permutations, the Kendall kernel is the difference between the number of elements that have the same order in both permutations and the number of elements that have different orders in both permutations, normalized by the number of pairs.

\subsubsection{Indefinite domain-independent kernels}
\label{subsec:kernel-functions:indefinite}

As our method is supposed to be used in a large variety of domains, we will present here a number of kernel methods that can be used with any distance function. However, these kernel functions are indefinite for a wide range of distances, including the ones used in our evaluation.

\vspace{1mm}
\noindent
\textbf{Distance substitution kernels:}
Haasdonk and Bahlmann \cite{Haasdonk2004} showed that given a distance function $\delta$ and an origin object $o$, one can induce a scalar product between these objects
\begin{equation}
\inner{o_a}{o_b}_\delta^o = \frac{1}{2} \left(\delta(o_a,o)^2 + \delta(o_b,o)^2 - \delta(o_a,o_b)^2\right)
\label{eq:dist_sub_prod}
\end{equation}
Using this scalar product, they proposed four kernel functions in object space
\begin{align*}
K_\delta^{lin} (o_a,o_b) &= \inner{o_a}{o_b}_\delta^o \\
K_\delta^{nd} (o_a,o_b) &= -\delta(o_a,o_b)^\beta, & \beta \in [0,2] \\
K_\delta^{pol} (o_a,o_b) &= \left(1 + \gamma \inner{o_a}{o_b}_\delta^o\right)^p, & \gamma \in \mathbb{R}^+, p \in \mathbb{N} \\
K_\delta^{rbf} (o_a,o_b) &= e^{-\gamma \delta(o_a,o_b)^2}, & \gamma \in \mathbb{R}^+
\end{align*}
Note that only weak assumptions are imposed on the distance function $\delta$: non-negative, symmetric, zero diagonal (i.e. $\delta(x,x)=0$). If some given distance function does not satisfy these requirements, it can easily be transformed to satisfy them. \corrected[2.14]{Concretely, it can be symmetrized by $\bar{\delta}(x, y) := \frac{1}{2} (\delta(x,y) + \delta(y,x))$, given zero diagonal by $\bar{\delta}(x,y) := \delta(x,y) - \frac{1}{2} (\delta(x,x) + \delta(y,y))$, and made positive by $\bar{\delta}(x,y) := \lvert\delta(x,y)\rvert$ \cite{Haasdonk2004}}. Thus, the distance substitution kernels above can be applied in conjunction with any distance function. However, it is only a positive definite kernel (or conditionally positive definite for $K_\delta^{nd}$) if the distance function is negative definite, i.e. isometric to an $L_2$-norm \corrected[2.15]{\cite{Haasdonk2004}}. As a large number of distance functions, for example the string edit distance, are not isometric to an $L_2$-norm, these kernels are often indefinite.

\vspace{1mm}
\noindent
\textbf{Combination kernel:}
Specifically for the domain of strings using the edit distance, Neuhaus and Bunke \cite{NEUHAUS20061852} proposed a combination of the above scalar product (Eq. (\ref{eq:dist_sub_prod})) using several origin objects $o_i \in \bar{O} \subset O$
\begin{align*}
K_S^+(o_a, o_b) =& \sum_{o_i \in \bar{O}} \inner{o_a}{o_b}_\delta^o\\
K_S^*(o_a, o_b) =& \prod_{o_i \in \bar{O}} \inner{o_a}{o_b}_\delta^o
\end{align*}
Using a subset of strings instead of only one origin object, they aim to reduce the dependence on this reference string. Note that no edit distance specific properties were used, and both kernels can be used with any distance function $\delta$ similar to the distance substitution kernels. From both combination methods we propose to use $K_\delta^{comb} = K_S^+$ due to its distance-preserving nature, as shown in section \ref{subsec:kernel-distance-preserving}.

\subsubsection{Distance-preserving kernels}
\label{subsec:kernel-distance-preserving}
Several of the above shown kernel functions are able to preserve the distance in the vector space of their respective projection. Although having distance-preserving positive definite kernels would be preferable, the above mentioned kernel functions $K^{ssk}$, $K^{part}$ and $K^{kend}$ are not. $K^{ssk}$ has no direct relation to the string edit distance at all, while for $K^{part}$ and $K^{kend}$ the distance is proportional to the squared norm in kernel space (a proof for $K^{kend}$ is given in \cite{jiao2017kendall}).

In the following we will show how a number of the previous shown indefinite kernels are distance-preserving for specific parameters.
Using Eq. (\ref{eq:kernel-norm}), one can show that $K_\delta^{comb}$ and $K_\delta^{lin}$ preserves distances in kernel space

\scalebox{0.9}{\parbox{\linewidth}{% parbox necessary to give line width to align
\begin{align*}
\lVert \kernel(o_a)-\kernel(o_b) \rVert_2 =& \left(K_\delta^{comb}(o_a,o_a) - 2 K_\delta^{comb}(o_a,o_b) \right. \\
& \left. + K_\delta^{comb}(o_b,o_b)\right)^\frac{1}{2} \\
=& \left(\sum_{o_i \in \bar{O}} \left(\frac{\delta(o_a,o_i)^2}{2}  + \frac{\delta(o_a,o_i)^2}{2} - \frac{\delta(o_a,o_a)^2}{2}  \right.\right. \\
&- \delta(o_a,o_i)^2 - \delta(o_b,o_i)^2 + \delta(o_a,o_b)^2 \\
&+ \left.\left. \frac{\delta(o_b,o_i)^2}{2}  + \frac{\delta(o_b,o_i)^2}{2}  - \frac{\delta(o_b,o_b)^2}{2}  \right)\right)^\frac{1}{2} \\
=& \left( \sum_{o_i \in \bar{O}} \delta(o_a,o_b)^2 \right)^\frac{1}{2} \\
=& \sqrt{|\bar{O}|} \  \delta(o_a,o_b)
\end{align*}
}}
In the special case of $|\bar{O}| = 1$, this is the $K_\delta^ {lin}$ kernel. For $\bar{O} > 1$, the distortion is constant which has no influence on the minimization of the median function (\ref{eq:median}), meaning that  it is an optimal distance-preserving embedding in kernel space. \corrected[2.16]{Similarly, it can be shown that using $K_\delta^{nd}$, the embedding leads to}
\[ \corrected{\lVert \kernel(o_a)-\kernel(o_b) \rVert_2 = \sqrt{2} \delta(o_a,o_b)^\frac{\beta}{2}} \]
\corrected{meaning that it is distance-preserving for $\beta=2$. For $K_\delta^{pol}$ using $p=1$ it leads to}
\[ \corrected{\lVert \kernel(o_a)-\kernel(o_b) \rVert_2 = \sqrt{\gamma} \delta(o_a,o_b)} \]
\corrected{and is therefore distance-preserving for $p=1$ and any $\gamma \neq 0$, in particular $\gamma = 1$.}

The observation above makes all four distance-preserving options ideal candidates for kernel-based median computation using the above mentioned parameters, since there are no distortions between distances in kernel space. This should lead to a better median computation in kernel space and ultimately better median reconstruction, which is indeed confirmed by the experimental results reported in Section \ref{sec:evaluation}.

\subsection{\corrected[2.2]{Complexity of the proposed framework}}
\label{subsec:complexity}

\begin{table}
    \caption{\corrected[2.2]{Complexity of the kernel-based median computation framework}}
    \label{tbl:complexity}
    \begin{tabular}{l|l}
         Algorithm steps & Complexity \\ \hline
         Precomputing kernel values $K(o_i,o_j)$ & $\mathcal{O}(n^2)$  \\
         Precomputing weights $\omega_i$ & $\mathcal{O}(j_{max} \cdot n^2)$\\\hline
         Linear/triangular reconstruction & $\mathcal{O}(n)$\\
         Linear/triangular recursive reconstruction & $\mathcal{O}(n \log n)$ \\\hline
    \end{tabular}
\end{table}

\corrected[2.2]{The asymptotic complexity of the proposed methods can be seen in Table \ref{tbl:complexity}. Assuming the weighted mean function and kernel function can be evaluated in constant time, one needs in total $O(j_{max} \cdot n^2)$ time for precomputing the kernel values and weights $\omega_i$, as these do not change in the reconstruction process. Note that $j_{max}$ is quite low in practice (see the evaluation in Section \ref{subsec:evaluation-convergence}). For linear and triangular reconstruction (Algorithm \ref{alg:kernel-linear}) additional $O(n)$ time is needed for evaluating $\alpha$ a constant number of times with $O(n)$ each. Similarly, $\mathcal{O}(n \log n)$ additional time is needed for linear and triangular recursive reconstruction (Algorithm \ref{alg:kernel-lin-rec}) by $O(\log n)$ evaluations of $\alpha$. Overall, our algorithm thus has the complexity $O(n^2)$.}

%\subsection{Notes on the generalized median approximation}
\subsection{Discussions}
\label{subsec:notes-approximation}

In the previous sections we have shown how the generalized median can be approximated using positive definite and indefinite kernel functions, and given a number of examples for kernel functions that could be used using any distance function. Although a few of approximations are made, among which are the transfer of $\alpha$ computed by the orthogonal projection in vector space onto an object space where there might not be such a clear projection, or the use of indefinite kernels where the Weiszfeld algorithm may not find a minimum or even converge, or the computation of $\alpha$ using complex weights whose interpretation is not clear. Thus, we will evaluate the convergence of the Weiszfeld algorithm using indefinite kernels in Section \ref{subsec:evaluation-convergence} and the computation of $\alpha$ in kernel space in Section \ref{subsec:evaluation-reconstruction} to show that our assumptions are reasonable.

\corrected[2.17]{In addition, it should be noted that using a kernel to compute the generalized median in kernel space can change the distance at hand. I.e. instead of minimizing the sum of distances, we minimize the sum of distances of embedded objects. Therefore, it is important to use kernel functions related to the distance in the original space.}

%All presented kernel functions are applicable to any domain and use the original distance formulation in their definition (see Section \ref{subsec:kernel-functions}), meaning that the generalized median approximation in kernel space is likely $\mathcal{NP}$-hard itself. Therefore it is reasonable that the generalized median framework introduces a number of approximations to the computation, even in cases that are mathematically not entirely clear.

\section{Evaluation}
\label{sec:evaluation}

In this section we present the experimental results using the method presented in Section \ref{sec:kernel-mean}. First, we introduce the datasets used in our study in Section \ref{subsec:evaluation-datasets}. 
In Section \ref{subsec:evaluation-distance-preservation} we will show how the previously discussed kernel functions preserve distances in kernel space, and give reasons for why they are expected to produce better results than other methods. Section \ref{subsec:evaluation-convergence} investigates the convergence of the Weiszfeld algorithm using indefinite kernels, while Section \ref{subsec:evaluation-reconstruction} studies the reconstruction of the median object using the weighted mean. Section \ref{subsec:evaluation-results} compares the results of the proposed kernel-based method to the previous Prototype \cite{ferrer2010generalized} and Distance-Preserving Embedding Frameworks \cite{nienkoetter2019dpe} with explicit vector space embedding. Then, we study the correlation of the median quality (in terms of sum of distances) and the degree of distance preservation in Section \ref{subsec:evaluation-distance-correlation}. Finally, a discussion of the computational time is presented in Section \ref{subsec:evaluation-time}.

\subsection{Datasets}
\label{subsec:evaluation-datasets}

Table \ref{tbl:datasets} shows the datasets used in our study. To ensure a wide range of applications we tested 6 datasets divided into three types, from which two were artificially generated and four real. The exact number of sets in each dataset and number of objects in each set can be found in the table.

The string datasets Darwin and CCD consist of character strings using the Levenshtein edit distance for the median computation. The Darwin dataset is artificially generated using lines of Charles Darwins famous work ``On the Origin of Species'' that were randomly modified by substitution, insertion and deletion of characters according to real-life OCR error rates \cite{jiang2012generalized}. The Copenhagen Chromosome Dataset (CCD) is a dataset containing encoded parts of chromosome sequences \cite{Lundsteen-al80}. In both cases the corresponding weighted mean function applies string edit operations from the Levenshtein edit distance until the ratio in Eq. (\ref{eq:weighted-mean}) is reached \cite{Bunke_2002}, \restructured{as seen in Section \ref{sec:reconstruction}.}

\begin{table}[t]
	\caption{Evaluated datasets for generalized median computation}
	\label{tbl:datasets}
	\scalebox{0.845}{
	\begin{tabular}{c|c|c|c|c}
		      Dataset        &      Type       & \#Sets &  \#Objects in  &      Distance      \\
		                     &                 &        &    each set    &      function      \\ \hline
		       Darwin        &     String      &   36   &       40       &    Levenshtein     \\
		        CCD          &     String      &   22   &      100       &    Levenshtein     \\
		    UCI Cluster      &     Cluster     &   8    &       25       & Partition Distance \\
		    Gen. Cluster     &     Cluster     &   8    &       20       & Partition Distance \\
		    ranking-bio      & Ranking w. Ties &   40   &       7        &  gen. Kendall-$\tau$   \\
		    ranking-real     & Ranking w. Ties &   40   & 8-17 (mean 12) &  gen. Kendall-$\tau$   \\ \hline
	\end{tabular}}
\end{table}

The UCI Cluster and Gen. Cluster datasets contain integer label vectors encoding clusterings of unknown data.  The UCI Cluster dataset was created by clustering data from the UCI Data Repository \cite{bache2013uci} with k-means clustering using varying parameters. Each set of the Gen. Cluster dataset consists of a random base label vector that was modified by random changes in labels using the method in \cite{franek2014ensemble}. Here, the Partition Distance was used, which counts the number of objects that have to swap their assigned cluster into a different one for both clusterings to become the same partition of the data. The labels of each cluster are disregarded. For the weighted mean function, we change $\mbox{round}(\alpha \cdot \delta(o_a,o_b))$ of disagreeing labels of the first cluster $o_a$ into the corresponding labels of the second cluster $o_b$ \cite{Franek_2014}.

The ranking-bio and ranking-real datasets are two real world datasets \cite{brancotte2015rank} and contain ranking with ties. The first one, ranking-bio, consists of the BioMedical dataset, which includes only a very small number of real world rankings in each set. The ranking-real dataset is composed of the F1, Ski Cross, Ski Jumping and WebSearch datasets. These sets consist of the results of different sport tournaments and the order of search results in different web search engines. In both cases the generalized Kendall-$\tau$ distance is used \cite{Boulakia_2011}. This distance function measures the number of disagreements between rankings. For example, if $x$ is ranked ahead of $y$ in the first ranking, but $y$ is ranked ahead in the second, it is counted as one disagreement, otherwise not. Since ties are a possibility, it is counted as 0.5 disagreements if $x$ and $y$ are tied in one ranking but not tied in the other. The distance is the summed number of disagreements between all ranked items. As weighted mean function, starting from $o_a$ we iteratively move ranked items with the largest number of disagreements into new positions with lower number of disagreements towards $o_b$, until the required ratio $\alpha$ is reached. \corrected[2.4]{Note that there exist simple methods such as the classical Borda and Copeland method for median computation in ranking space. It can be easily shown that these methods are the generalized median computation (\ref{eq:median}) based on simple distance functions. In our work we use the more sophisticated generalized Kendall-$\tau$ distance, which leads to a $\mathcal{NP}$-hard median computation problem \cite{Boulakia_2011}. Thus, our approach is particularly suitable to apply.}

It should be pointed out that these distance functions are not necessarily negative definite, as is the requirement of the distance substitution kernels to be positive definite. Using the modifications shown in Section \ref{subsec:gen-med-indefinite-kernel}, we can however still apply our method using these kernel functions to compute the generalized median in kernel space. As will be seen in the results, the kernel method is still able to compute a good approximation of the generalized median for such kernels.

\subsection{Distance preservation of kernel functions}
\label{subsec:evaluation-distance-preservation}

Since the generalized median computation considerably benefits from a good preservation of distances \cite{nienkoetter2019dpe}, we measured the degree of distance preservation of the kernels. For each kernel introduced in Section \ref{subsec:kernel-functions}, we computed all pairwise distances in kernel space using Eq. (\ref{eq:kernel-norm}), and determined the ratio $c$ between original distance and Euclidean distance in vector space
\begin{align}
\delta(o_i,o_j) &= c \cdot \lVert \kernel(o_i) - \kernel(o_j) \rVert_2
\label{eq:distance-relation}
\end{align}
A perfect distance-preserving embedding is achieved if there is a constant $c$ for all object pairs.

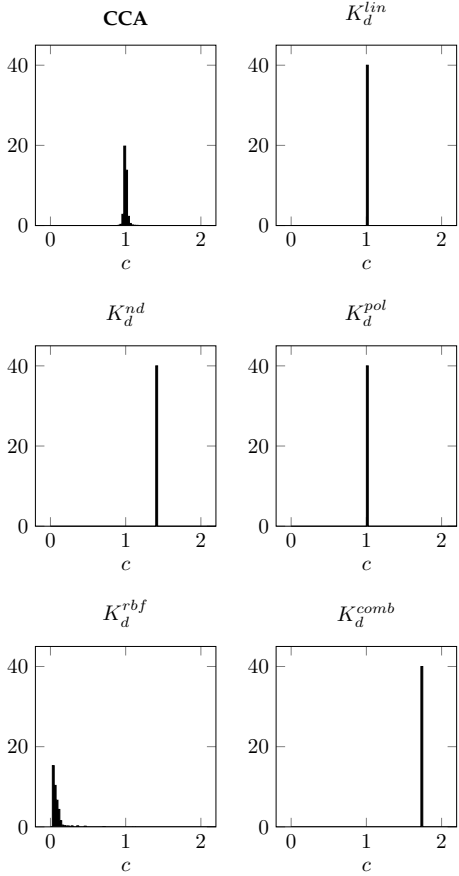
\begin{figure}
%	\scalebox{0.7}{\include{images/distortion/distortion_darwin}}
	%\pgfplotstableread{img/distortion/distortion_darwin.txt}{\datatable}
\pgfplotstableread[col sep=comma]{images/distortion/distortion_darwin.csv}{\datatable}
\def\ymax{45}
\def\xmax{2.2}
\def\axwidth{3cm}

\centering
\begin{tikzpicture}[scale=0.8]

% for alignment
%\node[at={(-0.5cm,0cm)}](n) {};

\begin{axis}[%
width=\axwidth,
height=3cm,
at={(0cm,0cm)},
scale only axis,
xmin=-0.2,
xmax=\xmax,
xlabel={$c$},
ymin=0,
ymax=\ymax,
ylabel={},
axis background/.style={fill=white},
title style={font=\bfseries},
title={CCA},
%extra x ticks={4.472}, extra x tick labels={$\sqrt{d}$},
]
%\addplot[ybar interval,black,fill=blue] table[x=c,y=prototype] from \datatable; 
\addplot[const plot,black,fill=black] table[x=c,y=cca] from \datatable; 
\end{axis}

\begin{axis}[%
width=\axwidth,
height=3cm,
at={(4cm,0cm)},
scale only axis,
xmin=-0.2,
xmax=\xmax,
xlabel={$c$},
ymin=0,
ymax=\ymax,
ylabel={},
axis background/.style={fill=white},
title style={font=\bfseries},
title={$K_d^{lin}$},
%extra x ticks={4.472}, extra x tick labels={$\sqrt{d}$},
]
\addplot[const plot,black,fill=black] table[x=c,y=dist_sub_lin] from \datatable; 
\end{axis} 

\begin{axis}[%
width=\axwidth,
height=3cm,
at={(0cm,-5cm)},
scale only axis,
xmin=-0.2,
xmax=\xmax,
xlabel={$c$},
ymin=0,
ymax=\ymax,
ylabel={},
axis background/.style={fill=white},
title style={font=\bfseries},
title={$K_d^{nd}$},
%extra x ticks={4.472}, extra x tick labels={$\sqrt{d}$},
]
\addplot[const plot,black,fill=black] table[x=c,y=dist_sub_nd] from \datatable; 
\end{axis} 

\begin{axis}[%
width=\axwidth,
height=3cm,
at={(4cm,-5cm)},
scale only axis,
xmin=-0.2,
xmax=\xmax,
xlabel={$c$},
ymin=0,
ymax=\ymax,
ylabel={},
axis background/.style={fill=white},
title style={font=\bfseries},
title={$K_d^{pol}$},
%extra x ticks={4.472}, extra x tick labels={$\sqrt{d}$},
]
\addplot[const plot,black,fill=black] table[x=c,y=dist_sub_pol] from \datatable; 
\end{axis} 

\begin{axis}[%
width=\axwidth,
height=3cm,
at={(0cm,-10cm)},
scale only axis,
xmin=-0.2,
xmax=\xmax,
xlabel={$c$},
ymin=0,
ymax=\ymax,
ylabel={},
axis background/.style={fill=white},
title style={font=\bfseries},
title={$K_d^{rbf}$},
%extra x ticks={4.472}, extra x tick labels={$\sqrt{d}$},
]
\addplot[const plot,black,fill=black] table[x=c,y=dist_sub_rbf] from \datatable; 
\end{axis} 

\begin{axis}[%
width=\axwidth,
height=3cm,
at={(4cm,-10cm)},
scale only axis,
xmin=-0.2,
xmax=\xmax,
xlabel={$c$},
ymin=0,
ymax=\ymax,
ylabel={},
axis background/.style={fill=white},
title style={font=\bfseries},
title={$K_d^{comb}$},
%extra x ticks={4.472}, extra x tick labels={$\sqrt{d}$},
]
\addplot[const plot,black,fill=black] table[x=c,y=edit] from \datatable; 
\end{axis} 
\end{tikzpicture}

%\end{axis}
%\end{tikzpicture}
	\caption{Histogram of the distance distortion constant $c$ for the Darwin dataset. The used parameters for kernel methods are $\beta = 2$ for $K_\delta^{nd}$, $\gamma = 1$ and $p = 1$ for $K_\delta^{pol}$, and $|\bar{O}| = 3$ for $K_\delta^{comb}$. Explicit embedding methods used $0.8 \cdot n$ dimensions where $n$ is the number of objects.}
	\label{fig:dist-darwin}
\end{figure}

For comparison, we used the embedding method CCA, which showed the best distance preservation for these datasets \cite{nienkoetter2019dpe}. The results for the Darwin dataset as an example can be seen in Figure \ref{fig:dist-darwin}. For all object sets and pairwise objects of this dataset these histograms show the distribution of the constant $c$ in Eq. (\ref{eq:distance-relation}). The more $c$ is located at one constant, the better the distances are preserved in the implicit embedding of the kernel or the explicit embedding of CCA. As theoretically shown in Section \ref{subsec:kernel-distance-preserving}, $K_\delta^{lin}$, $K_\delta^{nd}$, $K_\delta^{pol}$ and $K_\delta^{comb}$ achieve a perfect distance-preserving embedding. $K_\delta^{rbf}$ however does not guarantee a distance-preserving embedding and is shown to be slightly worse than CCA. Similar observation could also be made on the other five datasets. Thus, we expect that $K_\delta^{lin}$, $K_\delta^{nd}$, $K_\delta^{pol}$ and $K_\delta^{comb}$ will have the most accurate object representation in vector space and accordingly the best median result.

\subsection{Convergence of the Weiszfeld algorithm for indefinite kernels}
\label{subsec:evaluation-convergence}

As shown in Section \ref{subsec:kernel-functions}, many domain-independent kernels are not necessarily positive definite, leading to complex values in the computation of the Weiszfeld algorithm if used as is (Section \ref{subsec:gen-med-indefinite-kernel}). For vector spaces -- and therefore positive definite kernels -- the convergence of the Weiszfeld algorithm is well understood \cite{uster2000convergence}. However, these convergence proofs do not hold in the case of indefinite kernel functions. As we are unable to formally prove the convergence in this case, we conducted a simulation to test if the algorithm converges for a set of given objects using infinite kernels.

\begin{table}[t]
	\caption{\corrected[1.1]{}\corrected[1.2]{Convergence of the Weiszfeld algorithm for 5 indefinite kernel functions $K_\delta^{lin}$, $K_\delta^{nd}$, $K_\delta^{pol}$, $K_\delta^{rbf}$, $K_\delta^{comb}$ (first value in each column) and 3 positive definite kernel functions $K^{ssk}$, $K^{part}$, $K^{kend}$ (second value in each column).}}
	\label{tbl:convergence}
	\centering	
	\begin{tabular}{l|ccc}
	    Dataset (num sets) & max-iter & med-iter & complex weights\\ \hline
        Darwin (36)             & 29 / 13   & 9 / 11     & 28/0 (of 180) \\
        CCD (22)                & 17 / 33    & 12 / 24     & 0/0 (of 110) \\
        Gen-Cluster (8)         & 14 / 12    & 10 /10    & 0/0 (of 40) \\
        UCI-Cluster (8)         & 27 / 19    & 20 / 15    & 0/0 (of 40) \\
        ranking-bio (40)        & 144 / 85    & 42 / 76    & 2/0 (of 200) \\
        ranking-real (40)       & 31 /23     & 17 / 17    & 0/0 (of 200) \\ \hline
        random-string (1000)    & 11 / 11     & 10 / 10     & 0/0 (of 5000) \\ 
        random-cluster (1000)   & 14 / 11    & 12 / 10     & 0/0 (of 5000) \\
        random-ranking (1000)   & 8 / 9     & 6 / 8     & 0/0 (of 5000) \\ \hline
	\end{tabular}
\end{table}

In addition to the datasets shown in Section \ref{subsec:evaluation-datasets}, we generated 1000 sets of randomly generated strings, clusterings and rankings each for this test. Each randomly generated set consists of 50 to 150 objects of size 50 to 100. Here, the size is the number of characters for strings, the number of clustered objects for clusterings and the number of ranked values for rankings, respectively. \corrected[1.3]{For each string in the random-string dataset, random letters were uniformly drawn from all letters of the english alphabet. Each clustering in random-clustering was created by uniformly drawing random integer labels to fill a label vector. The number of clusters varies between 3 and 10. For rankings, a random permutation was created, again drawn uniformly from all possible permutations of a given length.}

For each set we computed the weights of the generalized median in kernel space using Equation (\ref{eq:kernel-weiszfeld1}) for all five presented (possibly) indefinite kernels $K_\delta^{lin}$, $K_\delta^{nd}$, $K_\delta^{pol}$, $K_\delta^{rbf}$ and $K_\delta^{comb}$  as well as three positive definite kernels $K^{ssk}$, $K^{part}$ and $K^{kend}$ (Section \ref{subsec:kernel-functions}).

Table \ref{tbl:convergence} shows the resulting statistics for our test. \corrected[1.1]{}\corrected[1.2]{For each dataset this table reports the maximum (median) number of iterations for convergence, max-iter (med-iter), over all tested sets, as well as the absolute amount of complex-valued weights $\omega_i$ (Eq. \ref{eq:kernel-weiszfeld1-complex}) that appeared over all datasets (complex weights). The first value in each column is computed only using indefinite kernels shown in Section \ref{subsec:kernel-functions:indefinite}, while the second value is computed using the domain-dependent positive definite kernel as described in Section \ref{subsec:kernel-functions:pos-definite}.} In all datasets using all kernels, the Weiszfeld algorithm in the form shown in Equation (\ref{eq:kernel-weiszfeld1}) converges in a relatively short amount of iterations, including the cases where complex values appeared. The maximum number of iterations (max-iter) was less than 150 in all cases, with much less than 30 in the majority of the datasets. The median number of iterations (med-iter) is less than 20 in the majority of datasets, showing that the Weiszfeld algorithm converges in a very short amount of time. \corrected[1.2]{The number of iterations for the indefinite kernels are in most cases only slightly larger than the values for positive definite kernels, showing that they converge equally well in practice.} In all tested datasets, complex weights only appeared in a total of 30 cases \corrected[1.2]{using indefinite kernels, and -- as expected -- nowhere for positive definite kernels}. This shows that even when using indefinite kernel functions, the weights remain real values in a large majority of the time.

In summary, the Weiszfeld algorithm converged in all cases we have tested, even though we cannot give a formal proof that it is always the case for indefinite kernels.

\subsection{Experiments on the generalized median reconstruction}
\label{subsec:evaluation-reconstruction}

As discussed in Section \ref{sec:history}, the weighted mean function can not guarantee that the computed approximated median is closer to the true generalized median than the previous objects. Nevertheless, one can assume that even if it is not always the case, using a large number of combinations leads to a better approximation than using a small number of weighted mean combinations. For this reason the linear recursive and triangular recursive reconstruction methods shown in Algorithm \ref{alg:kernel-lin-rec} will likely deliver a better approximation than linear and triangular methods shown in Algorithm \ref{alg:kernel-linear}.

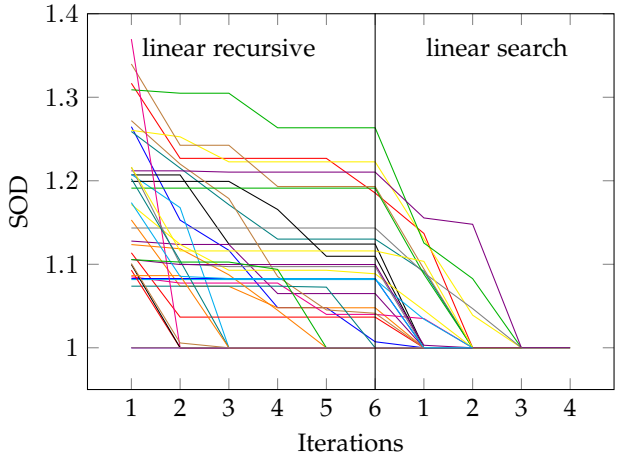
\begin{figure}
    \centering
    % absolute values
%\pgfplotstableread[col sep=comma]{images/sod_iteration.csv}{\datatable}
%\def\ymin{600}
%\def\ymax{1100}

% normalized values
\pgfplotstableread[col sep=comma]{images/sod_iteration_normalized.csv}{\datatable}
\def\ymin{0.95}
\def\ymax{1.4}

% non-normalized
\begin{tikzpicture}
\begin{axis}[%
    width=7cm,
    height=5cm,
    at={(0cm,14cm)},
    scale only axis,
    xlabel={Iterations},
    ymin=\ymin,
    ymax=\ymax,
    %axis equal,
    xtick={1,...,10},
    xticklabels={1,2,3,4,5,6,1,2,3,4},
    ylabel={SOD},
    axis background/.style={fill=white},
    cycle list name=color list,
    ]
    
    %\pgfmathsetmacro\wavelen{400+(700-400)*15/36}
    %\definecolor{mywavecolor}{wave}{100}
    %\colorlet{mycolor}[rgb]{mywavecolor}

    \foreach \y in {1, ..., 36} {
        %\definecolor{mycolor}{rgb}{\y,0,0}
        \addplot+ table[x=x,y=y\y] from \datatable;
    }

    \addplot+[mark=none,black] coordinates {(6, -1) (6, 1.5)};
    
    \node at (axis cs: 3,1.36) {linear recursive};
    \node at (axis cs: 8.5,1.36) {linear search};
\end{axis}
\end{tikzpicture}
    \caption{Minimization of the SOD on the Darwin dataset using linear-recursive (first 6 iterations) and linear search (second 4 iterations). Values are normalized by the final SOD.}
    \label{fig:sod_iteration}
\end{figure}

Figure \ref{fig:sod_iteration} shows the evolution of the sum of distances in the reconstruction using linear recursive reconstruction (Algorithm \ref{alg:kernel-lin-rec}), followed by linear search (see Section \ref{subsec:evaluation-results}) on the Darwin dataset (a similar evolution was observed on the other datasets). \correctedfinal[3]{Each of these 36 lines (not all differential due to overlapping) represents one string set being minimized by these methods.} All values in one reconstruction process have been normalized by their final sum of distances. As can be seen, even though the notion of projection is not entirely clear in the space of strings, the linear recursive reconstruction iteratively improves the computed median by combining strings with the weighted mean function, leading overall to a lower sum of distances over time. One iteration in this sense is the pairwise combination of all objects in the current set (Lines 2 to 11 in Algorithm \ref{alg:kernel-lin-rec}). Linear search further improves this approximated median by performing a line search using the weighted mean between the computed approximated median and every object in the set. Note that linear search stops early when no further improvement could be made. \corrected[2.19]{Note that a monotone falling sum of distances of the best median object is ensured here and in all results of our evaluation by simply saving the previously best object and including it in the set of objects for the next iteration.}

As can be seen, both methods successfully improve the sum of distances of the approximated median over time, even though it is not guaranteed to improve in every iteration due to the often not-unique nature of the weighted mean for objects. Nevertheless, there is a significant enough improvement in a wide range of datasets such that these methods can be used to reconstruct an approximate generalized median.

\subsection{Median quality}
\label{subsec:evaluation-results}

Our kernel-based method was tested on the datasets presented in \ref{subsec:evaluation-datasets}. In all cases, we compared our method using the appropriate distances with the traditional embedding method (either prototype or CCA embedding). CCA was shown \cite{nienkoetter2019dpe} to consistently provide the best results and thus chosen as a representative for this comparison. As in \cite{nienkoetter2019dpe}, all explicit embedding methods use $0.8 \cdot n$ dimensions for the target vectors. The parameters of the domain-independent kernel functions are $\beta = 2$ for $K_\delta^{nd}$, $\gamma = 1$ and $p = 1$ for $K_\delta^{pol}$. For $K_\delta^{comb}$, $|\bar{O}|=3$ objects were chosen using a k-medians algorithm. These parameters were chosen to guarantee a distance-preserving embedding if possible, as discussed in Section \ref{subsec:kernel-distance-preserving}.

For each dataset and method, the average of sum of distances results over all sets within the respective dataset is reported in Table \ref{tbl:results}, with the best result of each reconstruction method marked in bold. Note that the sixth row in each table, e.g. $K^{ssk}$ for the Darwin dataset, is related to a kernel specific to the particular domain and thus varies in each case. Since the absolute SOD depends on the distance function and dataset, all values were normalized using the linear transform $\frac{x-LB}{LB}$, where LB is the lower bound of the generalized median computed with a linear programming method \cite{jiang2002optimal}. Using this, a result of 0 would be a result that is guaranteed to be an optimal generalized median, while a result of 1 would mean a result whose SOD is 100\% larger than the lower bound. Therefore, a lower result means a better approximated generalized median. Note, however, that the true generalized median does not necessarily have a result of 0 since the lower bound may not be reached. The true generalized median can be any value greater than 0. For comparison purpose Table \ref{tbl:normalization} shows the absolute values of the median SOD as well as the median of lower bounds (LB) over all sets of a dataset.

\corrected[1.5]{To further show the difference in median quality, we also measured if there is a statistically significant improvement using the presented kernel methods over the distance-preserving framework. Values marked with a plus (+) are significantly better than the previously best CCA embedding, while values marked with a star (*) have no significant difference to CCA according to the Wilcoxon signed rank test \cite{wilcoxon1945} with a p-value threshold of 0.05. The signed rank test measures if the distribution of differences in the results of two methods on the same data has a mean of zero or not. If it is not zero, then one method consistently performs better than the other.}

In addition to the reconstruction methods discussed in Section \ref{subsec:gen-med-kernel}, we also included the linear search method presented in \cite{nienkoetter2016reconstruction}. This method improves given approximations of the generalized median by a local search using the weighted mean between the approximated result and the objects in the set. As a post-processing, it is independent of the basic reconstruction method and can be used with kernel methods without modification. In our case, the linear recursive result was used as starting point.

As can be seen in Table \ref{tbl:results}, kernel methods consistently outperform or match the performance of explicit embedding methods. In most datasets and reconstructions, the four distance-preserving kernels $K_\delta^{lin}$, $K_\delta^{nd}$, $K_\delta^{pol}$ and $K_\delta^{comb}$ show the best results, being statistically better than CCA in 68 of 120 combinations of kernel and reconstruction methods, and equally good in 48 of 120 cases, leaving only four cases where CCA is statistically better than these kernels (ranking-bio using linear reconstruction). $K_\delta^{rbf}$, however, often fails to compute accurate median approximations.

Of the domain specific kernel functions, $K^{part}$ shows the best results, especially with being equally good or better than CCA in the UCI Cluster dataset. $K^{kend}$, however, shows results mostly worse than CCA with the exception of one case, and $K^{ssk}$ consistently underperforms for all string datasets. This is consistent with the fact that $K^{part}$ is the kernel function closest related to the distance function used in the domain, while $K^{kend}$ only shares a tangential relationship to the respective distance and $K^{ssk}$ which shares no similarity to the string edit distance.

This finding generally confirms the expectation awakened in Section \ref{subsec:evaluation-distance-preservation} that a perfect distance preservation should ensure more accurate median computation, even with a large amount of approximations in the computation. As all kernels with a perfect distance preservation perform nearly equally well in all datasets, one can assume that the specific kernel method is not as important to the median approximation as the fact that it is distance-preserving. In addition, they perform better with reconstruction methods involving more objects, like linear recursive or triangular recursive, where the correct relationships after the embedding play an even greater role in the reconstruction.

\begin{table}	
	\caption{Median Quality on Six Datasets: Comparison with SOD. \corrected[1.5]{Values marked by a + (*) are statistically significantly better than (similar to) CCA.}}
	\label{tbl:results}
	\raggedright
{\bf (a) Darwin dataset:}
\vspace{2mm}

\newcommand{\resulttablescalefactor}{0.95}

\centering
\scalebox{\resulttablescalefactor}{%
\begin{tabular}{l|l|l|l|l|l}
& linear & triangular & lin-rec & triang-rec & lin-search \\ \hline
$K_\delta^{lin}$ & \textbf{0.3683*} & \textbf{0.3090*} & \textbf{0.2070+} & \textbf{0.1969*} & 0.1056* \\
$K_\delta^{nd}$ & \textbf{0.3683+} & \textbf{0.3090*} & \textbf{0.2070+} & \textbf{0.1969*} & 0.1056* \\
$K_\delta^{pol}$ & \textbf{0.3683*} & \textbf{0.3090*} & \textbf{0.2070+} & \textbf{0.1969*} & 0.1056* \\
$K_\delta^{rbf}$ & 0.8284 & 0.8284 & 0.8284 & 0.8284 & 0.1925 \\
$K_\delta^{comb}$ & \textbf{0.3683+} & \textbf{0.3090*} & \textbf{0.2070+} & 0.2041* & 0.1056* \\
$K^{ssk}$ & 0.4107 & 0.3682 & 0.4023 & 0.2932 & 0.1487 \\ \hline
CCA & 0.3691 & 0.3144 & 0.2425 & 0.1988 & \textbf{0.0988} \\
Prototype & 0.6411 & 0.5705 & 0.2357* & 0.2912 & 0.1275 \\ \hline
\end{tabular}}

\vspace{5mm}

\raggedright
{\bf (b) CCD dataset:}
\vspace{2mm}

\centering
\scalebox{\resulttablescalefactor}{%
\begin{tabular}{l|l|l|l|l|l}
& linear & triangular & lin-rec & triang-rec & lin-search \\ \hline
$K_\delta^{lin}$ & \textbf{0.3103*} & \textbf{0.3024*} & \textbf{0.2693+} & \textbf{0.2769*} & \textbf{0.2330+} \\
$K_\delta^{nd}$ & \textbf{0.3103*} & \textbf{0.3024*} & \textbf{0.2693+} & \textbf{0.2769*} & \textbf{0.2330+} \\
$K_\delta^{pol}$ & \textbf{0.3103*} & \textbf{0.3024*} & \textbf{0.2693+} & \textbf{0.2769*} & \textbf{0.2330+} \\
$K_\delta^{rbf}$ & 0.6376 & 0.6376 & 0.6376 & 0.6376 & 0.2948 \\
$K_\delta^{comb}$ & \textbf{0.3103*} & \textbf{0.3024*} & \textbf{0.2693+} & \textbf{0.2769*} & \textbf{0.2330+} \\
$K^{ssk}$ & 0.4513 & 0.4455 & 0.4829 & 0.4560 & 0.2970 \\ \hline
CCA & 0.3177 & 0.3177 & 0.2812 & 0.2835 & 0.2590 \\
Prototype & 0.4693 & 0.4241 & 0.3001 & 0.2944 & 0.2452* \\ \hline
\end{tabular}}
\vspace{5mm}

\raggedright
{\bf (c) Gen. Cluster dataset:}
\vspace{2mm}

\centering
\scalebox{\resulttablescalefactor}{%
\begin{tabular}{l|l|l|l|l|l}
& linear & triangular & lin-rec & triang-rec & lin-search \\ \hline
$K_\delta^{lin}$ & \textbf{0.4575*} & \textbf{0.4567+} & \textbf{0.4412+} & 0.4567+ & \textbf{0.4075*} \\
$K_\delta^{nd}$ & \textbf{0.4575*} & \textbf{0.4567+} & \textbf{0.4412+} & 0.4567* & 0.4098* \\
$K_\delta^{pol}$ & \textbf{0.4575*} & \textbf{0.4567+} & \textbf{0.4412+} & 0.4567+ & \textbf{0.4075*} \\
$K_\delta^{rbf}$ & 0.6968 & 0.6968 & 0.6968 & 0.6968 & 0.4560* \\
$K_\delta^{comb}$ & \textbf{0.4575*} & \textbf{0.4567*} & \textbf{0.4412+} & \textbf{0.4428+} & 0.4098* \\
$K^{part}$ & 0.4583 & 0.4583 & \textbf{0.4412*} & 0.4583 & 0.4098* \\ \hline
CCA & \textbf{0.4575} & \textbf{0.4567} & 0.4575 & 0.4543 & 0.4183 \\
Prototype & 0.6055 & 0.5706 & 0.4966 & 0.4862 & 0.4176* \\ \hline
\end{tabular}}

\vspace{5mm}
\raggedright
{\bf (d) UCI Cluster dataset:}
\vspace{2mm}

\centering
\scalebox{\resulttablescalefactor}{%
\begin{tabular}{l|l|l|l|l|l}
& linear & triangular & lin-rec & triang-rec & lin-search \\ \hline
$K_\delta^{lin}$ & \textbf{0.3055+} & \textbf{0.3055+} & \textbf{0.3055*} & \textbf{0.3055+} & \textbf{0.2844*} \\
$K_\delta^{nd}$ & \textbf{0.3055+} & \textbf{0.3055+} & \textbf{0.3055*} & \textbf{0.3055+} & \textbf{0.2844*} \\
$K_\delta^{pol}$ & \textbf{0.3055+} & \textbf{0.3055+} & \textbf{0.3055*} & \textbf{0.3055+} & \textbf{0.2844*} \\
$K_\delta^{rbf}$ & 1.4435 & 1.4435 & 1.4435 & 1.4435 & 0.2907* \\
$K_\delta^{comb}$ & \textbf{0.3055+} & \textbf{0.3055+} & \textbf{0.3055*} & \textbf{0.3055+} & \textbf{0.2844*} \\
$K^{part}$ & \textbf{0.3055*} & \textbf{0.3055*} & \textbf{0.3055*} & \textbf{0.3055*} & \textbf{0.2844+} \\ \hline
CCA & 0.3078 & 0.3078 & 0.3078 & 0.3078 & 0.2929 \\
Prototype & 0.4385 & 0.4301 & 0.3321 & 0.3427 & 0.2964* \\ \hline
\end{tabular}}

\vspace{5mm}

\raggedright
{\bf (e) ranking-bio dataset:}
\vspace{2mm}

\centering
\scalebox{\resulttablescalefactor}{%
\begin{tabular}{l|l|l|l|l|l}
& linear & triangular & lin-rec & triang-rec & lin-search \\ \hline
$K_\delta^{lin}$ & \textbf{0.1113} & \textbf{0.1113+} & \textbf{0.1113+} & \textbf{0.1113+} & \textbf{0.1058+} \\
$K_\delta^{nd}$ & \textbf{0.1113} & \textbf{0.1113+} & \textbf{0.1113+} & \textbf{0.1113+} & \textbf{0.1058+} \\
$K_\delta^{pol}$ & \textbf{0.1113} & \textbf{0.1113+} & \textbf{0.1113+} & \textbf{0.1113+} & \textbf{0.1058+} \\
$K_\delta^{rbf}$ & 0.1458 & 0.1458 & 0.1458 & 0.1458 & 0.1324 \\
$K_\delta^{comb}$ & \textbf{0.1113} & \textbf{0.1113+} & \textbf{0.1113+} & \textbf{0.1113+} & \textbf{0.1058+} \\
$K^{kend}$ & 0.1187 & 0.1187 & 0.1412 & 0.1187 & 0.1063 \\ \hline
CCA & \textbf{0.1113} & \textbf{0.1113} & \textbf{0.1113} & \textbf{0.1113} & 0.1087 \\
Prototype & \textbf{0.1113} & \textbf{0.1113} & \textbf{0.1113} & \textbf{0.1113} & 0.1087 \\ \hline
\end{tabular}}

\vspace{5mm}

\raggedright
{\bf (f) ranking-real dataset:}
\vspace{2mm}

\centering
\scalebox{\resulttablescalefactor}{%
\begin{tabular}{l|l|l|l|l|l}
& linear & triangular & lin-rec & triang-rec & lin-search \\ \hline
$K_\delta^{lin}$ & \textbf{0.2818+} & 0.2818* & \textbf{0.2649+} & \textbf{0.2698+} & \textbf{0.2200+} \\
$K_\delta^{nd}$ & \textbf{0.2818+} & 0.2818* & \textbf{0.2649+} & \textbf{0.2698+} & \textbf{0.2200+} \\
$K_\delta^{pol}$ & \textbf{0.2818+} & 0.2818* & \textbf{0.2649+} & \textbf{0.2698+} & \textbf{0.2200+} \\
$K_\delta^{rbf}$ & 0.5647 & 0.5647 & 0.5647 & 0.5647 & 0.2662 \\
$K_\delta^{comb}$ & \textbf{0.2818+} & 0.2818* & \textbf{0.2649+} & 0.2743+ & \textbf{0.2200+} \\
$K^{kend}$ & 0.3058 & 0.3058 & 0.2783 & 0.3058 & 0.2240* \\ \hline
CCA & 0.2968 & \textbf{0.2812} & 0.2723 & 0.2831 & 0.2216 \\
Prototype & 0.3841 & 0.3826 & 0.3000 & 0.3203 & 0.2363 \\ \hline
\end{tabular}}
\end{table}

\begin{table}	
	\caption{Values Used for the Normalization of the Results}
	\label{tbl:normalization}
	\centering	
	\begin{tabular}{l|c|c}
         & median SOD & median LB \\ \hline
        Darwin & 755.00 & 619.50 \\
        CCD & 1094.00 & 861.00 \\
        Gen-Cluster & 891.00 & 606.50 \\
        UCI-Cluster & 2823.00 & 2033.50 \\
        ranking-bio & 2574.00 & 2304.00 \\
        ranking-real & 2398.50 & 1850.00 \\
        \hline
	\end{tabular}
\end{table}	

\begin{table}	
	\caption{Median Quality on Six Datasets: Comparison with Ranking}
	\label{tbl:ranking}
	\raggedright
{\bf (a) without domain-specific methods:}
\vspace{2mm}
	
\centering
\scalebox{0.9}{%
\begin{tabular}{l|l|l|l|l|l|l}
& linear & triangular & lin-rec & triang-rec & lin-search & total \\ \hline
$K_\delta^{lin}$ & 1.12 & \textbf{1.20} & 1.38 & \textbf{1.37} & \textbf{1.55} & \textbf{1.32} \\
$K_\delta^{nd}$ & \textbf{1.11} & \textbf{1.20} & \textbf{1.35} & 1.41 & 1.58 & 1.33 \\
$K_\delta^{pol}$ & 1.12 & \textbf{1.20} & 1.38 & \textbf{1.37} & \textbf{1.55} & \textbf{1.32} \\
$K_\delta^{rbf}$ & 6.23 & 6.27 & 6.42 & 6.41 & 5.07 & 6.08 \\
$K_\delta^{comb}$ & 1.15 & 1.24 & 1.39 & 1.40 & 1.58 & 1.35 \\ \hline
CCA & 1.42 & 1.99 & 2.73 & 2.61 & 2.58 & 2.27 \\
Prototype & 5.04 & 5.19 & 4.18 & 4.54 & 3.77 & 4.54 \\ \hline
\end{tabular}}
\vspace{5mm}

\raggedright
{\bf (b) with domain-specific methods:}
\vspace{2mm}

\centering
\scalebox{0.9}{%
\begin{tabular}{l|l|l|l|l|l|l}
& linear & triangular & lin-rec & triang-rec & lin-search & total \\ \hline
$K_\delta^{lin}$ & 1.18 & 1.27 & 1.43 & \textbf{1.42} & \textbf{1.72} & \textbf{1.40} \\
$K_\delta^{nd}$ & \textbf{1.18} & \textbf{1.26} & \textbf{1.39} & 1.45 & 1.74 & 1.40 \\
$K_\delta^{pol}$ & 1.18 & 1.27 & 1.43 & \textbf{1.42} & \textbf{1.72} & \textbf{1.40} \\
$K_\delta^{rbf}$ & 7.07 & 7.11 & 7.25 & 7.25 & 5.67 & 6.87 \\
$K_\delta^{comb}$ & 1.22 & 1.30 & 1.43 & 1.44 & 1.74 & 1.43 \\
$K^{ssk}$ & 5.30 & 5.25 & 6.37 & 6.21 & 4.95 & 5.62 \\
$K^{part}$ & 2.42 & 2.67 & 2.90 & 2.77 & 2.55 & 2.66 \\
$K^{kend}$ & 3.37 & 4.09 & 4.60 & 4.39 & 3.25 & 3.94 \\ \hline
CCA & 1.51 & 2.05 & 2.81 & 2.66 & 2.78 & 2.36 \\
Prototype & 5.68 & 5.79 & 4.51 & 4.92 & 4.13 & 5.01 \\ \hline
\end{tabular}}
\end{table}

In Table \ref{tbl:ranking} the average rank of each method is listed. Here we ranked the absolute sum of distances results of each method for each dataset and domain for a reconstruction method using 1 for the best result, 2 for the second best, and so on, meaning that a method with continually better results than other methods will have a lower rank. Additionally, the last column shows the average rank of each kernel and explicit embedding method over all reconstruction methods. $K_\delta^{lin}$ and $K_\delta^{pol}$ consistently show the best rank for each reconstruction method with $K_\delta^{nd}$ being slightly better for linear and triangular reconstruction but worse for linear-recursive, triangular-recursive and linear search. With a rank of around 1.3 to 1.4 in total, $K_\delta^{nd}$ and $K_\delta^{pol}$ are consistently on the first place for each dataset, meaning that it is a good choice for median computation in general. 
This confirms the discussion made in section \ref{subsec:evaluation-distance-preservation}, that is, a better distance approximation in vector space leads to a better median reconstruction. Comparing the domain specific kernel functions, it can also be seen that a closer relationship of the kernel function to the used distance leads to a better median approximation. $K^{part}$ and $K^{kend}$, while only being tangentially related to the respective distance functions, deliver a better result on average than $K^{ssk}$ which has no relationship to the string edit distance at all.

\subsection{Correlation with distance preservation}
\label{subsec:evaluation-distance-correlation}

\begin{figure}
	\scalebox{0.7}{\input{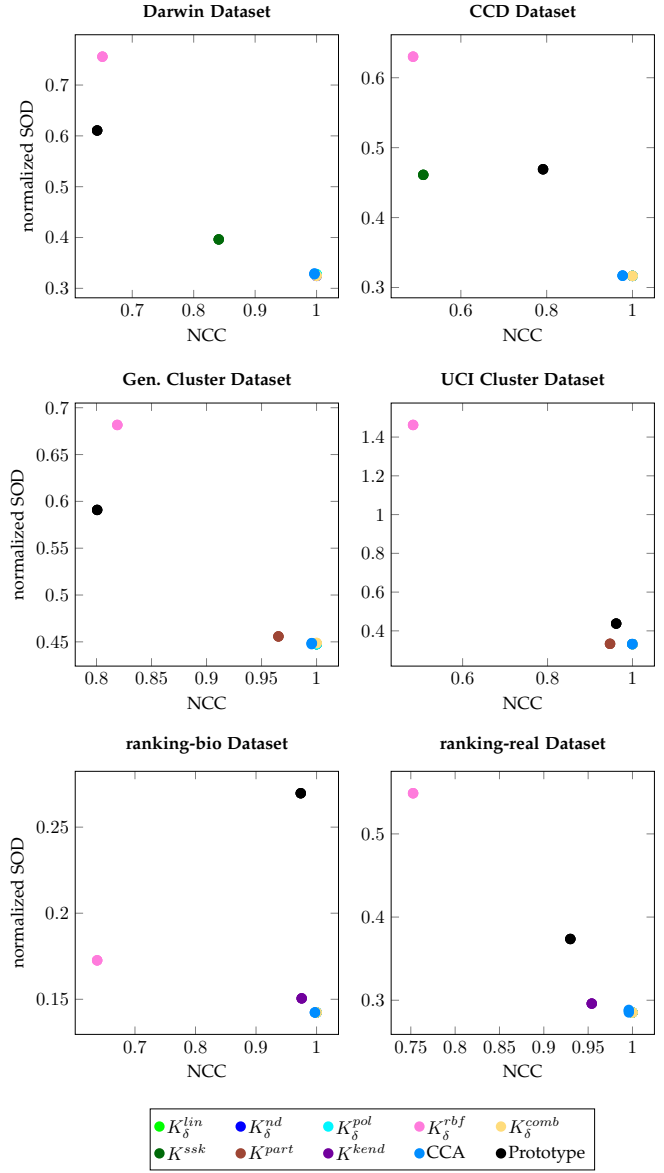}}
	\caption{Relationship between embedding and median quality in the different datasets, measured in NCC and SOD respectively.}
	\label{fig:ncc-sod}
\end{figure}

The relationship between our results and the distance-preserving properties of the kernels can be seen in Figure \ref{fig:ncc-sod}. In this figure, the Normalized Cross Correlation (NCC) between the distances in kernel space and the original distances is compared to the normalized sum of distances (SOD) result. The NCC measures how linearly dependent two sets of values are, and in this case, how closely Eq. (\ref{eq:distance-relation}) is approximated by one $c$. In a perfect embedding, the NCC would be 1, while it would be 0 if there is no linear relationship between the original distances and the ones in kernel or embedding space. As can be seen in Figure \ref{fig:ncc-sod}, a high NCC correlates with a good approximation of the generalized median, although with a few exceptions. For example in the String datasets, the prototype embedding is worse than $K_\delta^{rbf}$, even with a much higher NCC value.
	
\subsection{Computational time}
\label{subsec:evaluation-time}

\begin{table}
	\caption{Mean Computational Time on the Darwin Dataset (in Seconds)} 
	\label{tbl:time}
	\centering
\scalebox{1}{%
\begin{tabular}{l|l|l|l|l|l}
& linear & triangular & lin-rec & triang-rec & lin-search \\ \hline
$K_\delta^{lin}$ & 0.09 & 0.15 & 2.79 & 2.83 & 49.54 \\
$K_\delta^{nd}$ & 0.08 & 0.15 & 2.57 & 2.70 & 49.38 \\
$K_\delta^{pol}$ & 0.10 & 0.17 & 2.84 & 2.74 & 49.74 \\
$K_\delta^{rbf}$ & 0.08 & 0.14 & 2.59 & 2.66 & 78.98 \\
$K_\delta^{comb}$ & 0.11 & 0.18 & 3.08 & 3.06 & 49.88 \\
$K^{ssk}$ & 0.17 & 0.24 & 2.90 & 2.91 & 64.17 \\ \hline
CCA & 4.32 & 4.38 & 6.95 & 6.57 & 57.20 \\
Prototype & 0.07 & 0.13 & 2.74 & 2.61 & 58.09 \\ \hline
\end{tabular}}
\end{table}

All experiments were performed using a Intel Core i5-4590 (4 x 3.3GHz) with 16GB RAM using Python 3.5.
Table \ref{tbl:time} shows the average time of the median computation depending on the different reconstruction methods.
As expected, the more objects are involved in the reconstruction, the more time is needed, independent of if kernel methods or traditional embedding methods are used. As most of the time in the computation is used in the reconstruction, one can see that -- with the exception of linear search -- the run time is near constant for each embedding variant. Note that CCA has a near constant base-time of approximate 4 seconds due to the computational time of the much more complex initial embedding in each reconstruction method. Although computing the Weiszfeld algorithm using kernel functions is more complex than using embedded vectors, the fast convergence only causes very little additional time compared to Prototype embedding using the original Weiszfeld algorithm.
For linear search however, the methods display a larger variance in computational time. This can be explained by the nature of this reconstruction method. In contrast to the previous ones, linear search uses the result of linear-recursive as starting point and stops early once convergence is reached, meaning that due to the better base results, $K_\delta^{lin}$, $K_\delta^{nd}$, $K_\delta^{pol}$ and  $K_\delta^{comb}$ require less iterations of linear search than $K_\delta^{rbf}$, $K^{ssk}$, CCA and Prototype. As such, using  $K_\delta^{lin}$, $K_\delta^{nd}$, $K_\delta^{pol}$ and  $K_\delta^{comb}$ not only improves the embedding quality, but also the required time for the linear-search reconstruction. In particular, $K_\delta^{rbf}$ initially shows the worst sum of distances results by a large margin due to the high distortion of distances, which leads to a much higher required run-time of linear search.

In summary, we expect $K_\delta^{lin}$, $K_\delta^{nd}$, $K_\delta^{pol}$ and  $K_\delta^{comb}$ to have a lower computational time than CCA in all cases. All of these kernel methods can be easily computed from the distance function at hand, leading to fast kernel computations that preserve distances in the embedded space. CCA requires all pairwise distances for its embedding as well, but additional time for the computation of embedding vectors. Especially for large datasets and non-trivial embeddings, this can require a large number of iterations that are not present in distance based kernel methods. In contrast to the difference in embedding, the kernel Weiszfeld method is only barely more computationally complex as the vector-based Weiszfeld algorithm.
In combination with the sum of distances results, it is therefore recommended to use distance preserving kernel methods for median approximation instead of explicit embedding methods like CCA or Prototype.

%%%%%%%%%%%%%%%%%%%%%%%%%%%%%%%%%%%%%%%%%%%%%%%%%%%%%%%%%%%%%%%%%%%%%%%%%%%%%
\section{Toolbox}
\label{sec:toolbox}

We have implemented a toolbox in Python that contains a large number of embedding-based methods for consensus learning, including the most recent development of 
distance-preserving embedding method \cite{nienkoetter2019dpe} and our current work of kernel-based method. The user can apply these techniques on any dataset, provided a distance function and weighted mean function between two objects of the set are made available. It includes, among others, a variety of embedding, kernel, and reconstruction methods. It is implemented in such a modular manner that the user can easily insert additional methods and functions. In addition to embedding-based methods, the toolbox also contains other methods for generalized median computation such as the evolutionary weighted mean based
framework \cite{Franek_SSPR_2012}. In the toolbox part of the functions is also available in MATLAB.
By providing the toolbox for public use (available at: http://pria.uni-muenster.de), we encourage other researchers to explore generalized median computation and applications, 
especially for those problem instances with high inherent computational complexity.

%%%%%%%%%%%%%%%%%%%%%%%%%%%%%%%%%%%%%%%%%%%%%%%%%%%%%%%%%%%%%%%%%%%%%%%%%%%%%
\section{Conclusion}
\label{sec:conclusion}

In this paper we have shown how the generalized median can be computed using kernel methods, without using an explicit embedding like in previous methods. Although positive definite kernels would be preferred for the implicit embedding into an Euclidean vector space, we showed how indefinite kernels can also be used as an implicit embedding into a pseudo-Euclidean vector space. On six datasets, regularly better results could be demonstrated than methods using explicit embedding. Kernel-based generalized median computation not only shows overall superior results, but also overcomes the inherent drawbacks of current generalized median computation using explicit transformation. From the studied kernels, distance-preserving kernels $K_\delta^{lin}$, $K_\delta^{nd}$, $K_\delta^{pol}$ and $K_\delta^{comb}$ have, as expected, the best results, confirming the correlation between distance preservation and generalized median result. One should therefore use one of these kernel functions as baseline for median computation in practice.

\corrected[2.6]{Our goal in this work is to introduce a way to compute the generalized median for arbitrary spaces (with the weak requirement of weighted mean) and to demonstrate its good performance. As such, it is difficult to give theoretical guarantees for the quality of the generalized median in general. In fact, research papers in the literature with theoretical guarantees are typically done for a specific space explicitly using concrete knowledge of the space. Instead, we demonstrated superior performance compared to the previous distance-preserving and prototype embedding methods that represent the state of the art methods. For the comparison we used common datasets as in the publications of the previous methods, which can thus be considered as benchmark datasets. Overall, this performance comparison with state of the art methods on benchmark datasets helps to alleviate the lacking theoretical guarantees.}

Although in our experiments the Weiszfeld algorithm converges in all cases, we were not be able to formally prove its convergence in the general case of indefinite kernels. This will be further studied in future.
Aside from the computation of the median of a set, the kernel methods in this work can also applied to a number of other problems for more accurate computation. One example is kernel k-means, where a set of vectors is clustered in a kernel space to allow for non-convex clusters in the k-means algorithm \cite{dhillon2004kernel}. One could use the median computation from this paper to allow not only clustering of non-vector data, but also using the median instead of the mean.

\section*{Acknowledgements}
We sincerely thank the reviewers and the handling associate editor for their valuable comments that have substantially reshaped our paper. We thank Florian Eilers for
%his insight into dealing with complex vector spaces for this work. 
the valuable discussion about complex vector spaces.
This work was partly supported by the Deutsche
Forschungsgemeinschaft (DFG) – CRC 1450 – 431460824
and the European Union’s Horizon 2020 research and innovation programme under the Marie Sklodowska-Curie grant agreement No 778602 Ultracept.

%%%%%%%%%%%%%%%%%%%%%%%%%%%%%%%%%%%%%%%%%%%%%%%%%%%%%%%%%%%%%%%%%%%%%%%%%%%%%
\bibliographystyle{IEEEtran}
%\bibliography{IEEEabrv,literature}
\bibliography{literature}


% bibliographies in new column
\newpage

\begin{IEEEbiography}[{\includegraphics[width=1in,height=1.25in,clip,keepaspectratio]{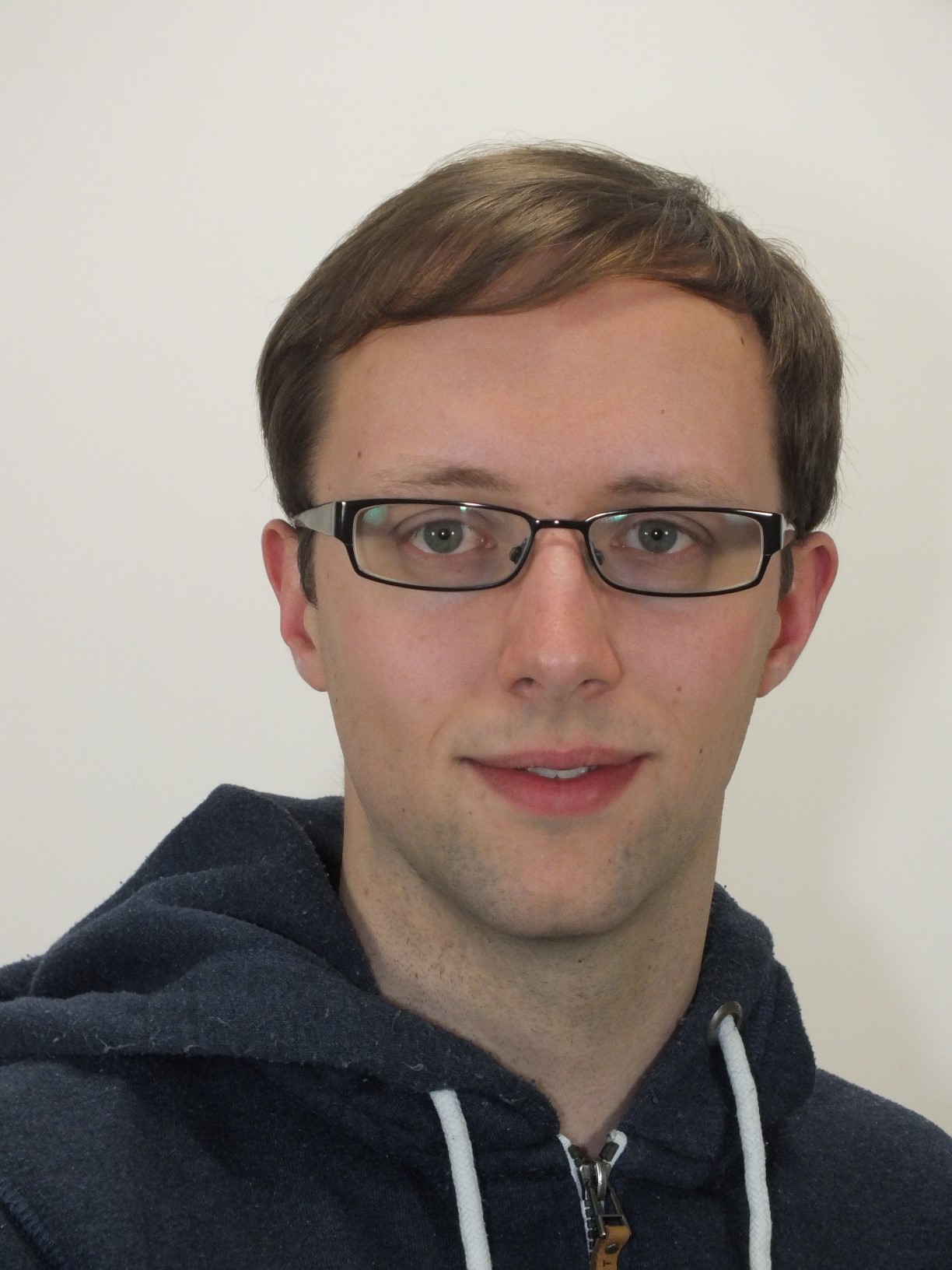}}]{Andreas Nienkötter}
received the master’s degree in computer science from University of Münster, Münster, Germany, in 2015, followed by a Ph.D. in 2021 at the same university. He is currently in a Postdoc position in Prof. X. Jiang’s Research Group for Pattern Recognition and Image Analysis. His research interests include consensus learning using the generalized median, vector space embedding methods, and dimensionality reduction methods.
\end{IEEEbiography}
\begin{IEEEbiography}[{\includegraphics[width=1in,height=1.25in,clip,keepaspectratio]{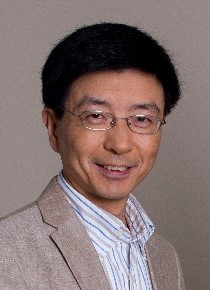}}]{Xiaoyi Jiang}
received the bachelor’s degree from Peking University, Beijing, China, and the Ph.D. and Venia Docendi (Habilitation) degrees
from the University of Bern, Bern, Switzerland, all in Computer Science. He was an Associate Professor with the Technical
University of Berlin, Berlin, Germany. Since 2002, he has been a Full Professor of Computer Science with the University of Münster, Münster, Germany, where he is currently the Dean of the Faculty of Mathematics and Computer Science. His current
research interests include biomedical imaging, 3D image analysis, and structural pattern recognition. Dr. Jiang is an Editor-in-Chief of International Journal of Pattern Recognition and Artificial Intelligence.
    He also serves on the Advisory Board and Editorial Board of several journals, including International Journal of Neural Systems and Journal of Big Data.
    Previously, he has been Associate Editor for
    IEEE Trans. on Systems, Man, and Cybernetics - Part B /  IEEE Trans. on Cybernetics, IEEE Trans. on Medical Imaging, and Pattern Recognition. He is Chair of IEEE EMBS Technical Committee on Biomedical Imaging and Image Processing (BIIP). He is  a Senior Member of IEEE and a Fellow of IAPR.
\end{IEEEbiography}



%\vfill


\end{document}